\documentclass[journal]{IEEEtran}
\ifCLASSINFOpdf
\else
\fi
\usepackage{times}
\usepackage{epsfig}
\usepackage{graphicx}
\usepackage{amsmath}
\usepackage{amssymb}
\usepackage{tabularx,booktabs}          % add by me
\usepackage{subfigure}   

\begin{document}
\title{Discriminative-Generative Dual Memory Video Anomaly Detection}

\author{Xin Guo, Zhongming Jin, Chong Chen, Helei Nie, Jianqiang Huang,
        Deng Cai, \IEEEmembership{Senior Member,~IEEE,} \\ Xiaofei He, \IEEEmembership{Senior Member,~IEEE,} and Xiansheng Hua, \IEEEmembership{Fellow, IEEE}% <-this % stops a space

\thanks{Xin Guo, Deng Cai, and Xiaofei He are with the State Key Lab of CAD\&CG, Zhejiang University, Hangzhou 310058, P.R.China  e-mail: guoxinzju@gmail.com, \{dengcai, xiaofeihe\}@cad.zju.edu.cn}

\thanks{Helei Nie is with Nanyang Technological University, Singapore 639798  e-mail: helei001@ntu.edu.sg}

\thanks{Zhongming Jin, Chong Chen, Jianqiang Huang, and Xian-Shen Hua are with the DAMO Academy, Alibaba Group, Hangzhou 311121, P.R.China  e-mail: \{zhongming.jinzm, cheung.cc, jianqiang.hjq\}@alibaba-inc.com, huaxiansheng@gmail.com}}

\maketitle

\begin{abstract}
Recently, people tried to use a few anomalies for video anomaly detection (VAD) instead of only normal data during the training process. A side effect of data imbalance occurs when a few abnormal data face a vast number of normal data. The latest VAD works use triplet loss or data re-sampling strategy to lessen this problem. However, there is still no elaborately designed structure for discriminative VAD with a few anomalies. In this paper, we propose a DiscRiminative-gEnerative duAl Memory (DREAM) anomaly detection model to take advantage of a few anomalies and solve data imbalance. We use two shallow discriminators to tighten the normal feature distribution boundary along with a generator for the next frame prediction. Further, we propose a dual memory module to obtain a sparse feature representation in both normality and abnormality space. As a result, DREAM not only solves the data imbalance problem but also learn a reasonable feature space. Further theoretical analysis shows that our DREAM also works for the unknown anomalies. Comparing with the previous methods on UCSD Ped1, UCSD Ped2, CUHK Avenue, and ShanghaiTech, our model outperforms all the baselines with no extra parameters. The ablation study demonstrates the effectiveness of our dual memory module and discriminative-generative network.
\end{abstract}


% \fi
\IEEEpeerreviewmaketitle

%%%%%%%%% BODY TEXT
\section{Introduction}
    Video anomaly detection (VAD) is to identify frames of anomalous events in a given video. It is widely used in video surveillance for public exception monitoring, traffic congestion, or real-time accident detection. However, anomaly detection is exceptionally challenging due to 1) Anomalies are unbounded. Any case that is far away from normal patterns is regarded as an anomaly. 2) The amounts of normal data and abnormal data are hugely imbalanced. Different from the vast number of normal daily events, abnormal events rarely happen in real-life situations.
    
    Almost all of the existing methods model anomaly detection in the video using \emph{non-anomaly assumption}, i.e., training a model describing normality without abnormal samples. The most popular VAD methods are based on reconstruction or prediction. Reconstruction based anomaly detection~\cite{lu2013abnormal,cong2011sparse, zhao2017spatio,hasan2016learning,luo2017revisit,gong2019memorizing} assumes that anomalies cannot be reconstructed accurately by a model learned only on normal data. Prediction based anomaly detection~\cite{zhao2017spatio,liu2018future,park2020learning}, on the other hand, uses the consecutive $t$ frames to predict the next frame and assumes anomaly will cause a large prediction error. However, the generalization ability of the model can also reconstruct some parts of an abnormal event, which makes the boundary of normality and abnormality unclear. Some articles~\cite{liu2019margin, ruff2019deep} propose introducing a few abnormal samples during training. Consequently, another challenge occurs along with a few anomalies in a vast amount of normal data: data imbalance. That is to say, the number of normal data is much larger than abnormal data. Liu et al.~\cite{liu2019margin} suggest using triplet loss to mitigate the imbalance issue. However, they do not elaborately design any structures that dedicate to capture the uniqueness of a few anomalies, and their model cannot make sure to work for unknown anomaly types.
    
    There are also many works to solve data imbalance problem~\cite{li2011semi, tian2011imbalanced, dong2017class, krawczyk2014cost, ren2018learning}. One may think using standard binary classification with some imbalance learning and data augmentation strategies should work pretty well. It is not the truth. In pure imbalance learning problems, the majority and the minority are equal in status except for the available data scale. It omits the particularity of anomaly detection that normalies are unbounded and any case that is far away from normal patterns can be regarded as an anomaly. It is inappropriate to simply regard the anomalies as one or several classes. As shown in paper~\cite{liu2019margin}, some methods (IVC with OS, IVC with OS \& FL, IVC with OS \& FL \& Two-stream, Triplet loss + OCSVM) based on imbalanced video classification methods are used as baselines. Those imbalance learing based can not outperform a special designed model for anomaly detection.
    
\begin{figure}[t]
    \centering
    \subfigure[Normality space of our model]{
        \label{fig:subfig:a_i} %% label for first subfigure
        \includegraphics[width=6.0cm]{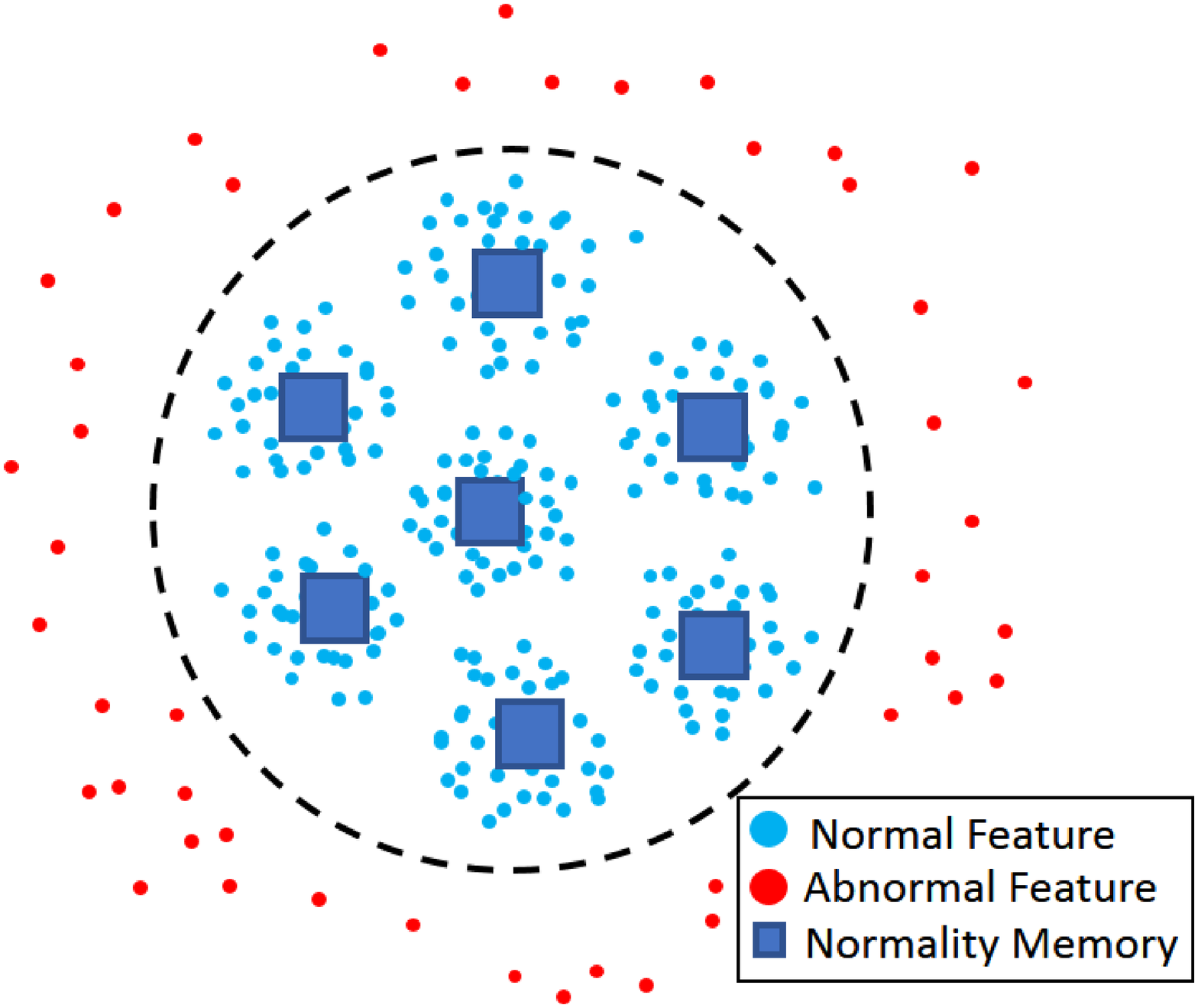}}
        \hspace{-0.2cm}
    \subfigure[Abnormality space of our model]{
        \label{fig:subfig:b_i} %% label for second subfigure
        \includegraphics[width=6.0cm]{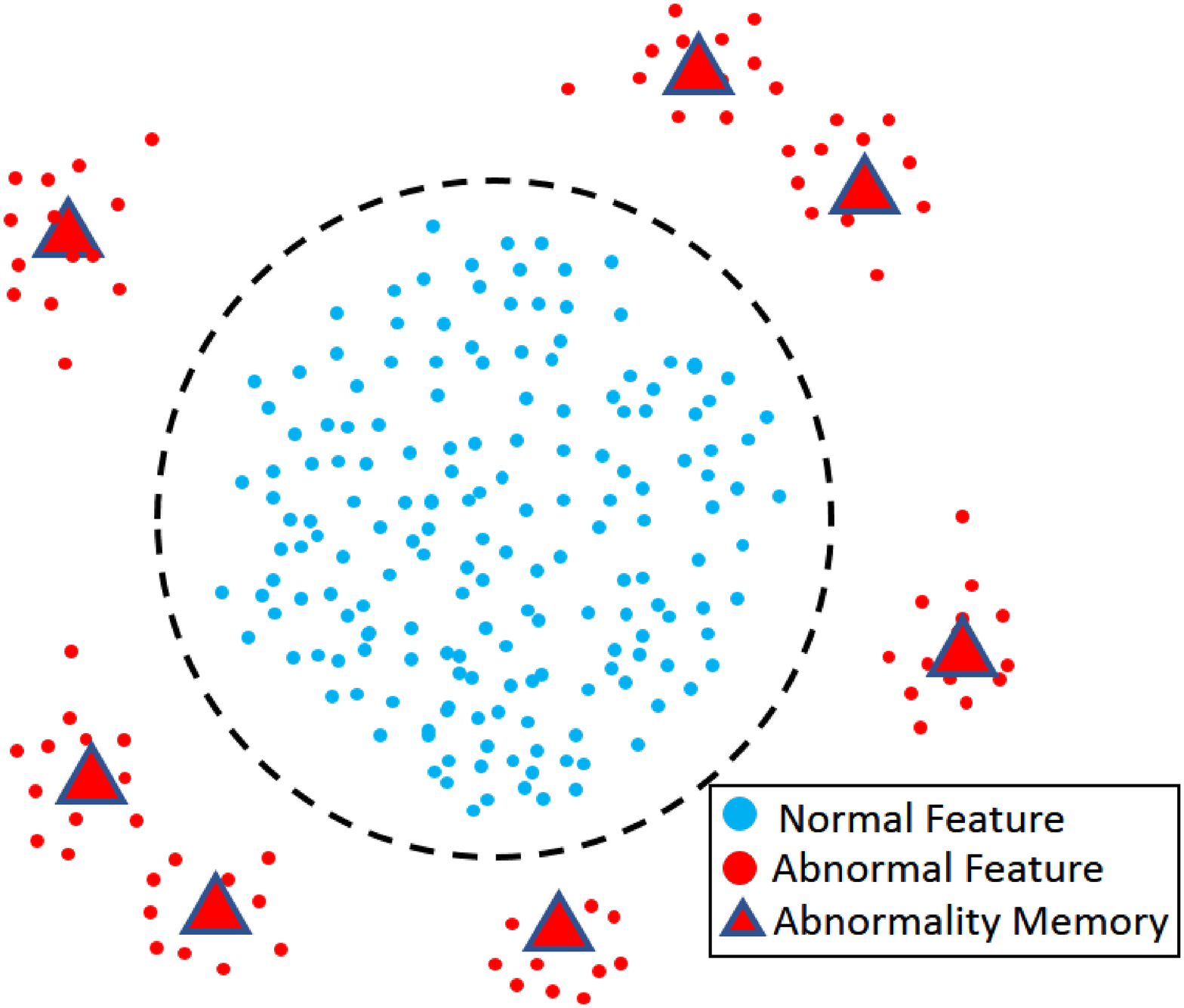}}
    \caption{A diagram for the feature space of our model. Normality/abnormality memories are used to record prototypical patterns of normal/abnormal data, respectively. To prevent abnormal samples from being overwhelmed by a vast number of normal ones, the normality and abnormality memory spaces are learned separately.}
    \label{fig:interpretation} %% label for entire figure
    \end{figure}

    In this paper, we propose a DiscRiminative-gEnerative duAl Memory (DREAM) model with two discriminators, a generator, and a dual memory module. The discriminators are used to tighten the normal data distribution boundary while the generator is used for frame prediction. Further, we propose a dual memory structure to separate the normality and abnormality memory space. As a result, the normal patterns are distributed in the blue region of normality memory space in Fig.~\ref{fig:subfig:a_i}. In this case, the abnormal samples can facilitate the learning of normal patterns to be more concrete and discriminative. For abnormality memory space in Fig.~\ref{fig:subfig:b_i}, similarly, normal samples can be introduced to render a more discriminative feature space for abnormal patterns. The final experiments show that our DREAM outperforms all the methods and that both the network design and the dual memory module benefit the final result.
    
    Our contribution can be summarized as follows: 
    
    1) \emph{We design a dual memory structure to address the data imbalance problem effectively. With our dual memory module, the normality and abnormality memories are updated separately so that the abnormal samples will not be overwhelmed by vest normal data. To our best knowledge, our DREAM is the first work that designs a particular module for video anomaly detection faced with data imbalance.}
    
    2) \emph{We use two discriminators and the memory update rule to learn a better embedding for the dual feature space. Besides memorize all of the high-frequent abnormal patterns, our DREAM also uses the abnormal data to force the model to get more reasonable feature space.}
    
    3) \emph{Our DREAM not only outperforms all of the models that use only normal samples but also exceeds the baselines that used abnormal samples but without the elaborately designed module. Further analysis shows that our model works for the unknown anomaly types.}
    
    In the following sections, we summarize the recent works related to video anomaly detection in Sec.~\ref{sec:related_work}. We introduce our model in Sec.~\ref{sec:our_model} and design several experiments to show superiority of DREAM in Sec.~\ref{sec:experiments}.

%------------------------------------------------------------------------
\section{Related work}
\label{sec:related_work}

    \textbf{Video Anomaly Detection} 
    This paper focuses on frame-level VAD. There are also recent works that focus on video-level week supervision~\cite{zaheer2020self,zaheer2020claws,wan2020weakly} and object-level detection~\cite{ionescu2019object,vu2019robust}, which beyond our discussion in this paper.
    
    \emph{Distance-based VAD} uses normal data to learn the feature and measures deviations from the normality to determine anomaly scores. One-class SVM~\cite{chen2001one} and the deep one-class network~\cite{ruff2018deep} are the classical one-class classification methods to tackle VAD problems. \cite{sabokrou2018adversarially} present one of the first approaches to use adversarial training for VAD. \cite{ramachandra2020street} propose a simple nearest-neighbor location-dependent anomaly detection scheme using hand-crafted along with hand-crafted distance measurement. Contrast~\cite{wang2020cluster} proposes to address the video anomaly detection task by a contrastive representation learning pretext task and introduce a novel instance discrimination pretext task, Cluster Attention Contrast.
    
    \emph{Probabilistic VAD} computes distance under a model in some probability space. Mahadevan et al.~\cite{mahadevan2010anomaly} propose learning a Mixture of Dynamic Textures (MDT) from training video patches, with the mixtures shared across larger “cell” regions. Cheng et al.~\cite{cheng2015video} propose a hierarchical local plus global method to detect anomalies. Scene-Aware~\cite{sun2020scene} builds a spatio-temporal context graph to model visual context information, including appearances of objects, spatio-temporal relationships among objects and scene types.
    
    \emph{GAN-based VAD} explores the possibility of improving the generative results using adversarial training. AbnormalGAN~\cite{ravanbakhsh2017abnormal} and Frame-Pred~\cite{gong2019memorizing} propose to use GAN for VAD, and their model is trained using normal frames and corresponding optical-flow images in order to learn an internal representation of the scene normality. OGNet~\cite{zaheer2020old} transforms the fundamental role of a discriminator from identifying real and fake data to distinguishing between good and bad quality reconstructions.
    
    \emph{Reconstruction-based VAD} assumes that a model trained on normal data will fail to properly reconstruct anomalies. \cite{lu2013abnormal,cong2011sparse,zhao2017spatio,luo2017revisit} propose to build sparse representation of the normal data. Hasan et al.~\cite{hasan2016learning} use fully convolutional AE as a reconstruction framework to detect anomalies. Gong et al.~\cite{gong2019memorizing} introduce autoencoder with memory block (MemAE) as a reconstruction model for anomaly detection. \cite{vu2019robust} propose to find unusual objects at high-level representations besides low-level data and to combine these detection results.
    
    \emph{Prediction-based VAD} uses the consecutive $t$ frames to predict the next frame and leverages the difference between a predicted future frame and its ground truth to detect an abnormal event. Zhao et al.~\cite{zhao2017spatio} perform 3-Dimensional convolution for spatial-temporal prediction. Liu et al.~\cite{liu2018future} train a frame-level prediction network by tracking object appearance and motion. The most related work to ours is Mem-Guided~\cite{park2020learning}. Mem-Guided~\cite{park2020learning} benefits from a specially designed memory structure for only normal data. We elaborately develop a dual memory module and a discriminative-generative network for both normal and abnormal data to solve data imbalance and feature learning.
    
    \textbf{Data Imbalance}
    The research on imbalanced classification can be generally divided into a data re-sampling strategy~\cite{li2011semi, tian2011imbalanced, dong2017class}, cost-sensitive learning~\cite{krawczyk2014cost, ren2018learning}. For data re-sampling strategy, Dong et al.~\cite{dong2017class} develop an end-to-end deep learning framework that is able to avoid the dominant effect of majority classes by using batch incremental hard sample mining of minority classes. For cost-sensitive learning, Ren et al.~\cite{ren2018learning} propose a novel meta-learning algorithm that learns to assign weights to the training examples based on their gradient directions. Huang et al.~\cite{huang2019deep} demonstrate that more discriminative deep representation can be learned by enforcing a deep network to maintain inter-cluster margins both within and between classes.

    \begin{figure*}
    \begin{center}
        \includegraphics[width=\linewidth]{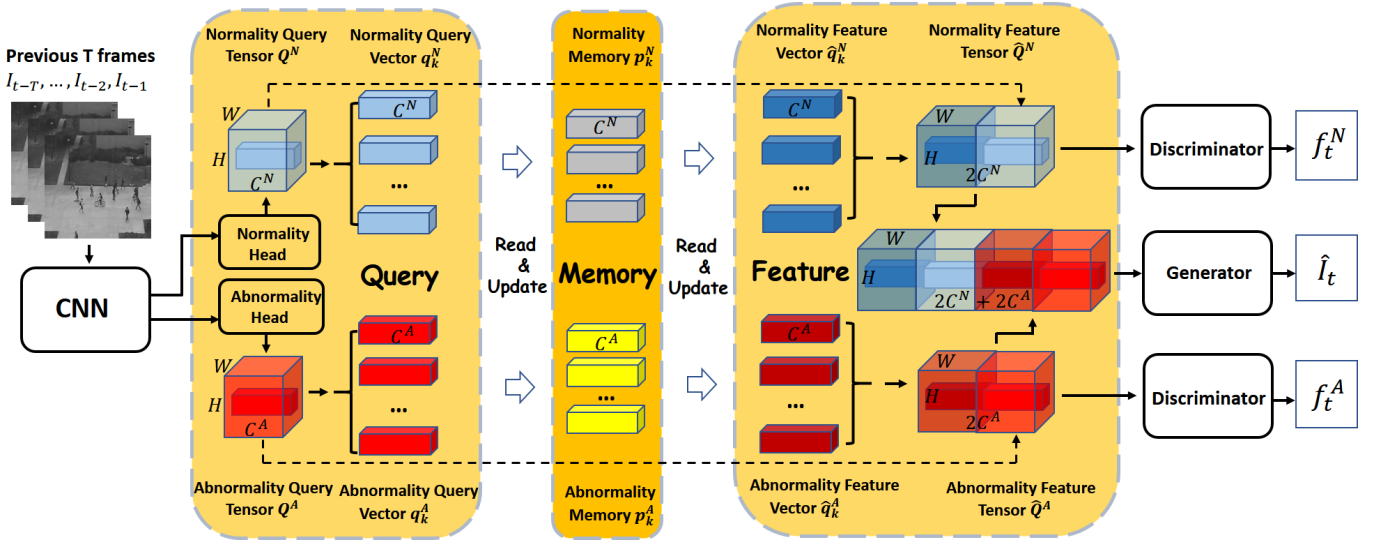}
    \end{center}
       \caption{Overview of DREAM. Both normal and abnormal data will go through the normality and abnormality branches, represented in blue and red. The low dimension normality and abnormality queries are extracted using a CNN feature extractor. Using read and update operations between queries and memories, we get low dimension features in both normality and abnormality space. Two discriminators are used to learn the discriminative feature spaces, and a generator is used to predict the next frame.}
    \label{fig:dgdm}
    \end{figure*}

\section{Our model}
\label{sec:our_model}
\subsection{Motivation and overview}
\label{sec:motivation}
    A basic idea in memory-based network~\cite{gong2019memorizing, park2020learning} is to represent each kind of normal features using their nearest memory. Memories are used to learn the typical patterns of feature. The abnormal features are expected to be far away from all the normality memories and thus can be distinguished from the normal ones. However, because the normal memories are randomly and evenly distributed over the whole feature space, it is hard to make sure that the abnormal feature has sufficient difference from the normality memories. Thus, abnormality scores of normal and abnormal samples may not be distinguished obviously. A direct way is to learn a more discriminative feature space. In the following, we use both the terms \emph{feature space} and \emph{memory space} to represent this low dimension feature space.
    
    Another problem is that abnormal patterns can be overwhelmed by normal patterns due to severe data imbalance. As shown in Fig.~\ref{fig:subfig:a_t}, although the normality and abnormality memories are synchronously updated in the same feature space during the training phase, the model seems to regard the very few abnormal samples as noise. In other words, the model is essentially trained by a large number of normal data because of the robustness of the neural network.
    
    To solve the above problems and get a more convincing feature distribution, we propose to learn the normality and abnormality space separately. The basic structure of DREAM in Fig.~\ref{fig:dgdm} is a memory-based~\cite{gong2019memorizing, park2020learning} frame prediction network. First, the low dimension normality and abnormality queries are extracted using a CNN feature extractor. We use UNet~\cite{ronneberger2015u} and ConvLSTM~\cite{liu2019margin} as our backbone and the details can be found in Sec.~\ref{sec:implementation_details}. Using read and update operations between queries and memories, we get low dimension features in both normality and abnormality space. Two discriminators are used to learn the discriminative feature spaces, and a generator is used to predict the next frame. Benefiting from the discriminative feature and the dual memory, which solving the 'overwhelming' problem, our model works well on imbalanced video anomaly detection.
    
    In the following sections, we introduce the dual memory module in Sec.~\ref{sec:dualmemory}, the discriminative-generative network structure in Sec.~\ref{sec:network}, and the loss function and abnormality score in Sec.~\ref{sec:lossscore}. Finally, we show an intuitive interpretation of our model in Sec.~\ref{sec:modelinterp}.

\subsection{Dual memory}
\label{sec:dualmemory}
    \begin{figure}[t]
    \centering
    \subfigure[Read]{
        \label{fig:subfig:a_r} %% label for first subfigure
        \includegraphics[width=0.6\linewidth]{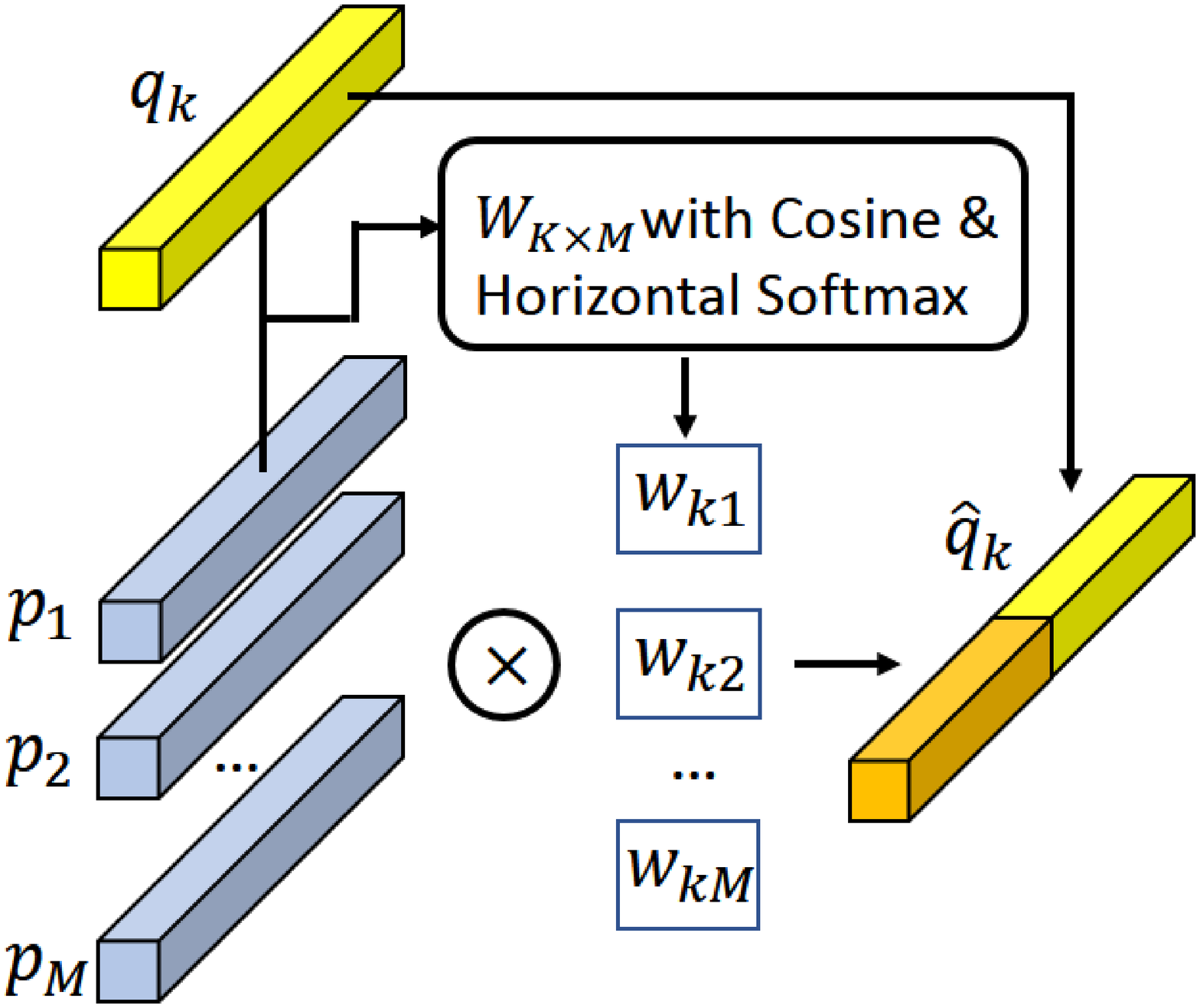}}
    \subfigure[Update normality space]{
        \label{fig:subfig:b_r} %% label for second subfigure
        \includegraphics[width=0.9\linewidth]{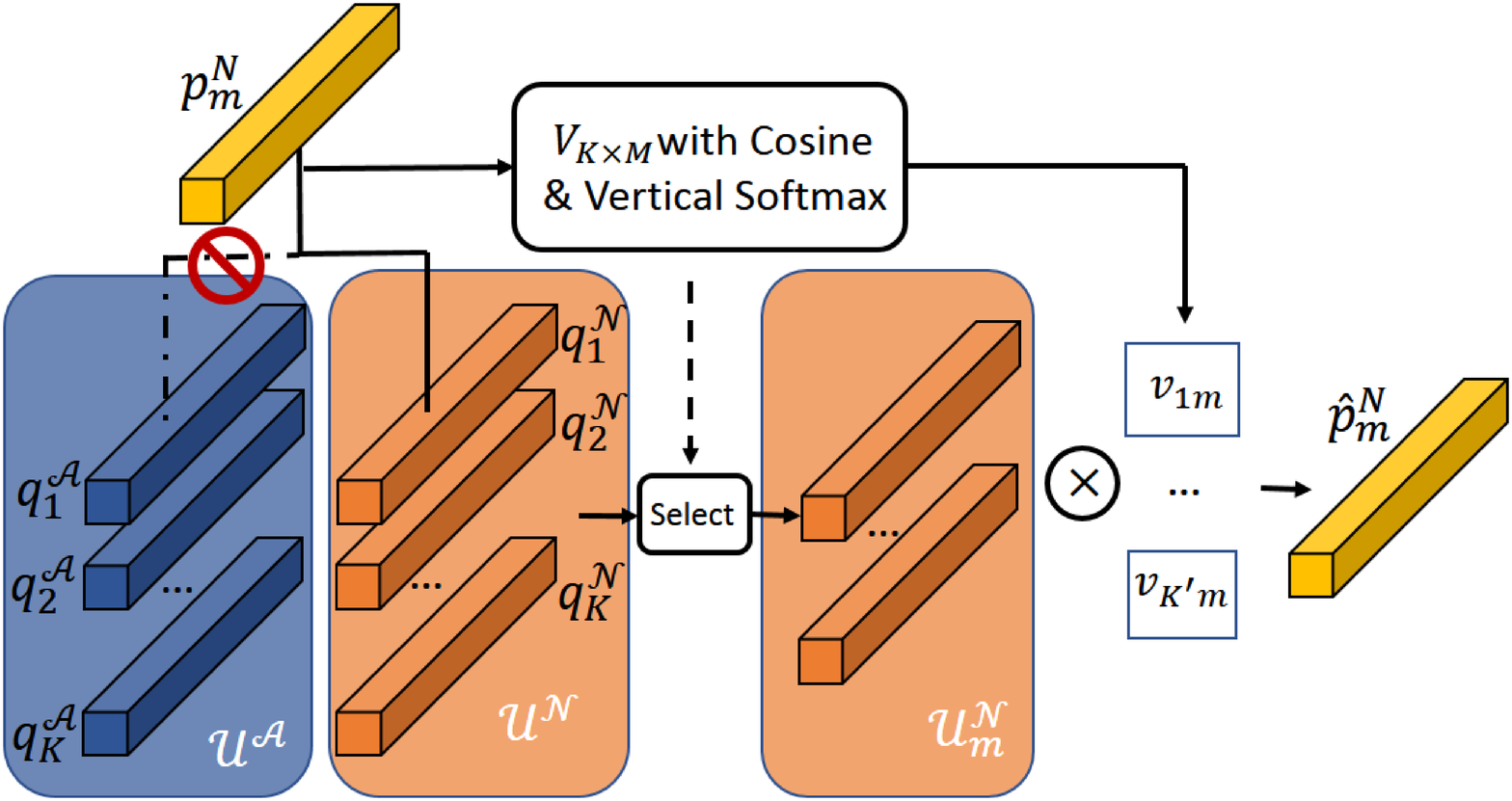}}
    \subfigure[Update abnormality space]{
        \label{fig:subfig:c_r} %% label for second subfigure
        \includegraphics[width=0.9\linewidth]{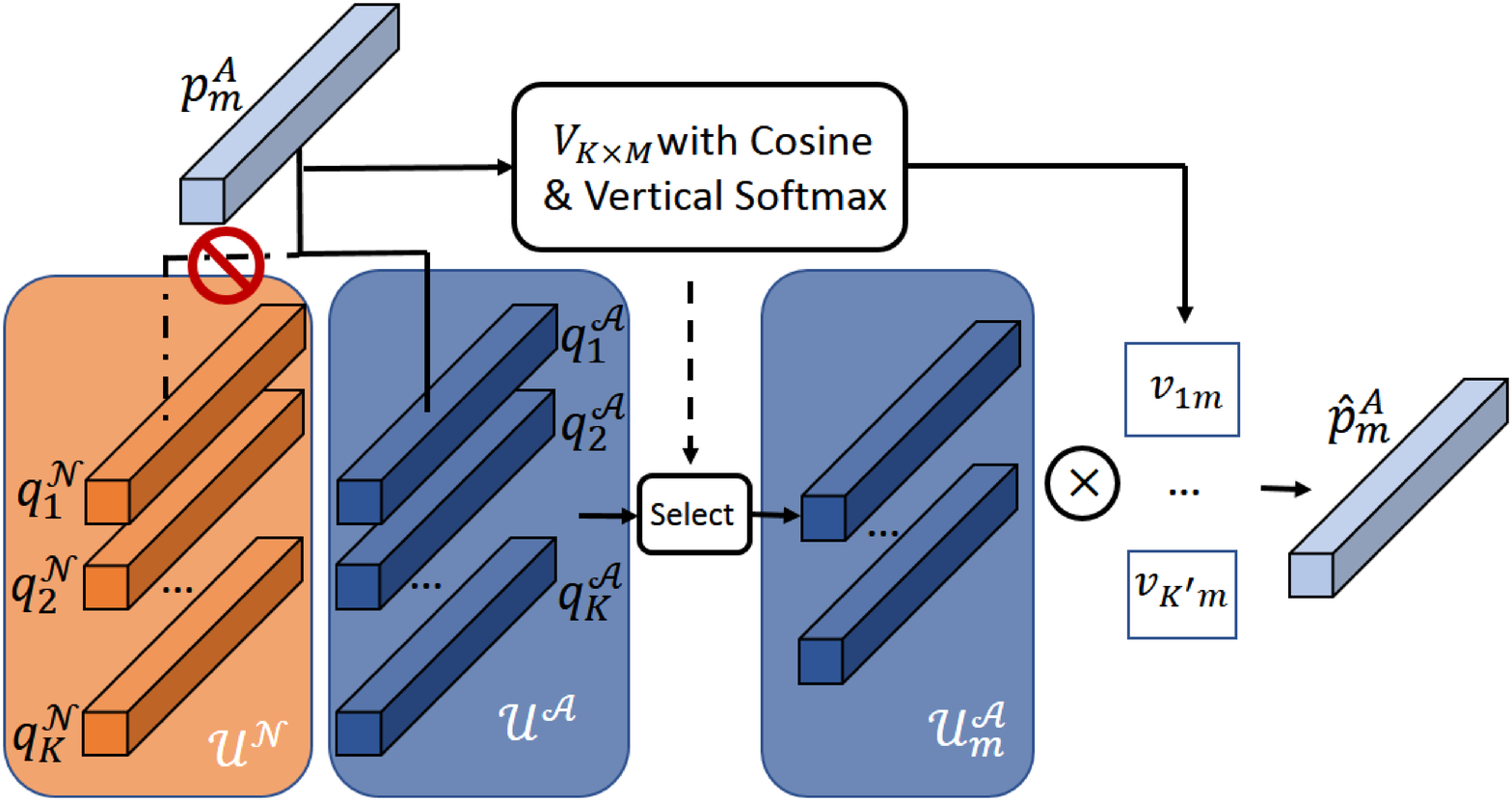}}
    \caption{Read and Update. (a) The normality and abnormality memory use the same read rule. (b) The normality memories are updated with only normal data. (c) The abnormality memories are updated with only abnormal data.}
    \label{fig:readupdate} %% label for entire figure
    \end{figure}

    A classical memory-based network consists of two terminologies: query $\mathbf{Q}$ and memory $\mathbf{P}$. 
    The \emph{query tensor} is the output of the CNN feature extractor and also the input of the dual memory module. We design two heads in our CNN feature extractor to output query tensors $\mathbf{Q}^{N}$, $\mathbf{Q}^{A} \in \mathbb{R}^{C \times H \times W}$ for normality and abnormality branches, respectively. $C$, $H$, $W$ are the number of channels, height, and width. The query tensor is then divided into $K (=H \times W)$ query vectors, and each vector is normalized to a unit vector on the $C$-dimension hypersphere. We use $\mathbf{q}_{k}^{N}$ and $\mathbf{q}_{k}^{A}$ to represent normality and abnormality query vectors, respectively. 
    
    There are two \emph{memory tensors} corresponding to the normality and abnormality query tensors, denoted by $\mathbf{P}^{N}$ and $\mathbf{P}^{A}$. Each memory tensor consists of $M$ memory vectors $\mathbf{p}_{m}\in{\mathbb{R}^{C}}$, which are sparse representations for the feature space. The vector number $M$ for normality and abnormality dictionary are not necessarily equal, and the memory vectors are also unit normalized.

    There are two basic operations for dual memory: to \emph{read} a new feature tensor using the input query and memory vectors and to \emph{update} the memory vectors using the input query vectors. We emphasize that \emph{both normal and abnormal data will pass through both the normality and abnormality branches in read operation}, and the normality query can be the query obtained from either a normal sample or an abnormal sample. The term \emph{normality} and \emph{abnormality} are used to distinguish the normality and abnormality branches, while \emph{normal} and \emph{abnormal} are used to represent the normal and abnormal samples. In this paper, we use the printing font superscripts $N$ and $A$ to represent the terms related to normality and abnormality branches, and the handwriting font superscripts $\mathcal{N}$ and $\mathcal{A}$ to represent the terms related to normal and abnormal data.

    \textbf{Read}: The read operation happens in both the training and testing phases and is showed in Fig.~\ref{fig:subfig:a_r}. We use the same rule for both the normality and abnormality branches, so we omit the superscripts $N$ and $A$ for brevity.
    
    For all the $K$ query vectors obtained from the query tensor, we calculate the cosine distance between each of the query vectors $\textbf{q}_{k}$ and each of the memory vectors $\mathbf{p}_{m}$. This results in a 2-dimensional correlation matrix $\mathbf{W}$ of size $K \times M$, and then a softmax operation is applied to $\mathbf{W}$ along the horizontal direction. We get the weight for $\mathbf{q}_{k}$ corresponding to all the memory vectors as Eq.~\eqref{equ:weight}.
    
    \begin{equation} \label{equ:weight}
        w_{km}= \frac{exp(\mathbf{q}_{k}^{T}\mathbf{p}_{m})}{\sum_{m'=1}^{M}exp(\mathbf{q}_{k}^{T}\mathbf{p}_{m'})}
    \end{equation}
    
    The output vector of the read operation is then calculated as a weighted average of weights $w_{km}$ with all memory vectors. All the $K$ new vectors are merged into a new feature tensor then concatenated with the initial query tensor along channel dimension to form $\hat{\mathbf{Q}} \in \mathbb{R}^{2C \times H \times W}$. The new feature vector $\hat{\mathbf{q}}_{k}$ is calculated as Eq.~\eqref{equ:read}.
    
    \begin{equation} \label{equ:read}
        \hat{\mathbf{q}}_{k}= concat(\sum_{m=1}^{M} w_{km} \mathbf{p}_{m}, \mathbf{q}_{k})
    \end{equation}
    
    After the read stage, the new normality feature tensor $\hat{\mathbf{Q}}^{N}$ and the new abnormality feature tensor $\hat{\mathbf{Q}}^{A}$ are concatenated to be a $2(C^{N}+C^{A}) \times H \times W$ tensor to form the input for the frame generator.

    \textbf{Update}: The update operation happens only in the training phase, and the rule for normality and abnormality memories are different. The update for normality and abnormality memory are showed in Fig.~\ref{fig:subfig:b_r} and \ref{fig:subfig:c_r} respectively.
    
    For the normality memories, we first collect the query $\textbf{q}_{k}^{\mathcal{N}}$ generated from the normality head with normal samples only. We denote this query set as $\mathcal{U}^{\mathcal{N}}$. Next, we calculate the cosine distance between each of items $\textbf{q}_{k}^{\mathcal{N}}$ in $\mathcal{U}^{\mathcal{N}}$ and each normality memory vectors $\mathbf{p}_{m}^{N}$ to obtain a 2-dimensional distance matrix $\mathbf{V}^{N}$. Unlike the read phase, this time, the softmax operation is applied to $\mathbf{V}^{N}$ along the vertical direction. According to the distance matrix, we collect all the normal query vectors that are considered as the nearest neighbors to the normality memory vector $\mathbf{p}_{m}^{N}$. We call them \emph{nearest normal queries from normality} for $\mathbf{p}_{m}^{N}$, represented by $\mathbf{q}_{k}^{\mathcal{N}} \in \mathcal{U}_{m}^{\mathcal{N}}$.
    
    For the normality memory vector $\mathbf{p}_{m}^{N}$, we update it with all the queries in $\mathcal{U}_{m}^{N}$ using Eq.~\eqref{equ:update1} as follows:
    
    \begin{equation} \label{equ:update1}
        \mathbf{p}_{m}^{N} \gets f(\mathbf{p}_{m}^{N} + \sum_{k \in \mathcal{U}_{m}^{\mathcal{N}}} v_{km} \mathbf{q}_{k}^{\mathcal{N}})
    \end{equation}
    in which, the $v_{km}$ is the vertical softmax of the distance matrix $\mathbf{V}^{N}$ and $f(\cdot)$ is the L2 normalization.

    Similar, the abnormality memory $\mathbf{p}_{m}^{A}$ is updated by queries $\textbf{q}_{k}^{\mathcal{A}}$ generated by abnormal samples. The queries generated from the abnormality head are collected as set $\mathcal{U}^{\mathcal{A}}$. The $\mathbf{V}^{A}$ and \emph{nearest normal queries from abnormality} $\mathcal{U}_{m}^{\mathcal{A}}$ for $\mathbf{p}_{m}^{A}$ are defined in similar way as normality memory update operations. For the abnormality memory word $\mathbf{p}_{m}^{A}$, we update it with all the queries in $\mathcal{U}_{m}^{A}$ using Eq.\eqref{equ:update2}.
    
    \begin{equation} \label{equ:update2}
        \mathbf{p}_{m}^{A} \gets f(\mathbf{p}_{m}^{A} + \sum_{k \in \mathcal{U}_{m}^{\mathcal{A}}} v_{km} \mathbf{q}_{k}^{\mathcal{A}})
    \end{equation}
    
    To summarize the dual memory mechanism: all normal and abnormal data will be feed into both normality and abnormality heads, producing query vectors for the read and update operations of the memory modules. The read stage will generate a new feature tensor $\hat{\mathbf{Q}}$ as the input for the two discriminators and the frame generator in the next stage. In the update stage, the normality memories are updated with only queries from the normal samples, while the abnormality memories are updated with only the queries from abnormal samples.

\subsection{Discriminative-generative network}
\label{sec:network}
    One of our intuition is that the feature of normal/abnormal data should be far away from the abnormality/normality memories. To achieve that, we use two discriminators with triplet loss~\cite{liu2019margin} to force normal data features to be close to each other, while the features of normal and abnormal data to be distant in both the normality and abnormality memory space.

    The principle of discriminator design is: the network should remain shallow to ensure the memory spaces can be sufficiently optimized by the loss. Therefore, we use a convolutional layer with $3 \times 3$ kernel for both of the feature tensor $\hat{\mathbf{Q}}^{N}$, $\hat{\mathbf{Q}}^{A}$ $\in \mathbb{R}^{2C \times H \times W}$, a global average pooling to further get the final input for the discriminative loss. 

\begin{figure}[t]
    \begin{center}
       \includegraphics[width=1.0\linewidth]{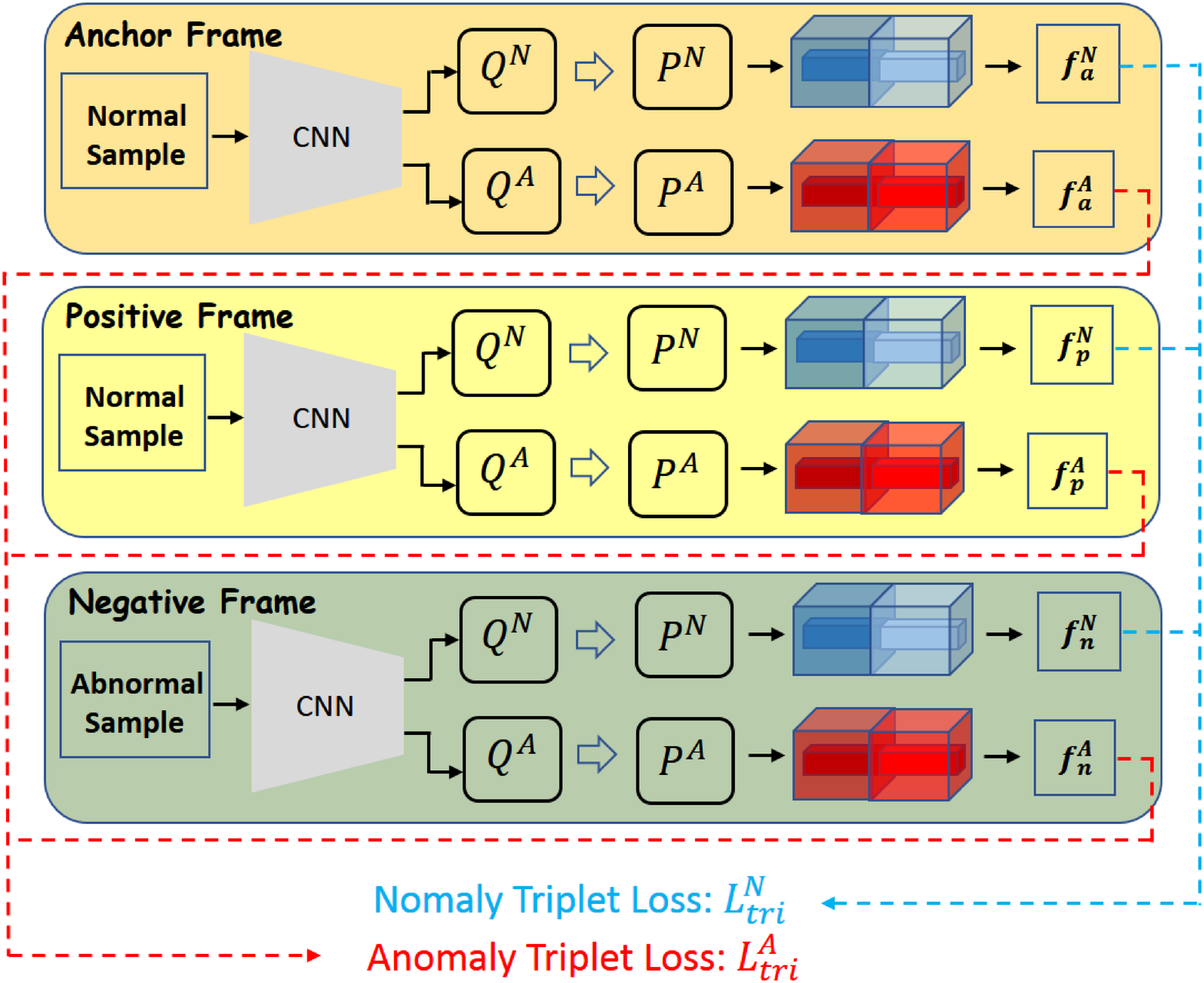}
    \end{center}
       \caption{Triplet loss. We use the normal data as the anchor as well as positive frames. We use the abnormal data as the negative frames. The triplet loss is used in both normality and abnormality branches.}
    \label{fig:tripletloss}
    \end{figure}

    Triplet loss can lead to a feature space with a one-class classification~\cite{chen2001one} formulation, thus we use it as the discriminative loss. For the output of the normality discriminative branch, we set the normal feature $f_{a}^{N}$, a randomly sampled normal feature $f_{p}^{N}$ and a randomly sampled abnormal feature $f_{n}^{N}$ as anchor, positive and negative samples, respectively. The triplet loss~\eqref{equ:triplet} will encourage a smaller distance between normal features, at the same time, encourage a larger distance between normal anchor and abnormal feature. As shown in Fig. \ref{fig:tripletloss}, it is used on both normality and abnormality discriminative branches.
    
    \begin{equation} \label{equ:triplet}
    \begin{split}
        L_{tri}(f_{a}^{N}, f_{p}^{N}, f_{n}^{N}) = &max(0, ||f_{a}^{N}-f_{p}^{N}||_{2}^{2}\\
        &-||f_{a}^{N}-f_{n}^{N}||_{2}^{2}+\beta)
    \end{split}
    \end{equation}
    
    To benefit from the frame prediction, we also use a generator to predict the next frame of the video sequence. We use the decoder in UNet~\cite{ronneberger2015u} or ConvLSTM~\cite{liu2019margin} as generator according to the backbone. We use a L2 distance between the predicted frame $\hat{\mathbf{I}}_{i}$ and the ground truth $\mathbf{I}_{t}$ as reconstruction loss in Eq.~\eqref{equ:reconstruction}. The generator and the reconstruction loss are necessary for video anomaly detection, because the area that is difficult to be reconstructed can be detected as the anomaly object. Only normal data are used in reconstruction loss, and thus the generator is trained using only normal data. 
    
    \begin{equation} \label{equ:reconstruction}
        L_{rec} = ||\mathbf{I}_{t} - \hat{\mathbf{I}}_{i}||_{2}^{2}
    \end{equation}
    
    Following the standard memory network~\cite{park2020learning}, we use the compactness loss and separateness loss to conduct a sparse effect for both feature space as well. The feature compactness loss~\eqref{equ:compactness} encourages the queries $\mathbf{q}_{k}$ to be close to the nearest item $\mathbf{p}_{p}$ in the memory while the feature separateness loss~\eqref{equ:separateness} encourages the query $\mathbf{q}_{k}$ and the second nearest item $\mathbf{p}_{n}$ to be distant, while the query and the nearest one to be nearby.

    \begin{equation} 
    \label{equ:compactness}
        L_{com} = \sum_{k}^{K}||\mathbf{q}_{k}-\mathbf{p}_{p}||_{2}^{2}
    \end{equation}
    \begin{equation} 
    \label{equ:separateness}
        L_{sep} = \sum_{k}^{K} max(0, ||\mathbf{q}_{k}-\mathbf{p}_{p}||_{2}^{2}-||\mathbf{q}_{k}-\mathbf{p}_{n}||_{2}^{2}+\alpha)
    \end{equation}

\subsection{Total loss and abnormality score}
\label{sec:lossscore}
    As shown in Eq.~\eqref{equ:finalloss}, the final loss function includes reconstruction loss $L_{rec}$ using only the normal samples, loss $L^N$ for the normality branch, and loss $L^A$ for the abnormality branch. $L^N$ consists of feature compactness loss $L_{com}^{N}$, feature separateness loss $L_{sep}^{N}$ and triplet loss $L_{tri}^{N}$. Meanwhile, $L^A$ consists of feature compactness loss $L_{com}^{A}$, feature separateness loss $L_{sep}^{A}$ and triplet loss $L_{tri}^{A}$. We use $\lambda$, $\mu$, and $\nu$ as weight parameters for the corresponding feature compactness loss, feature separateness loss, and triplet loss. % We omit the notation of the abnormality branch for brevity. 

    \begin{equation} \label{equ:finalloss}
    \begin{split}
        &Loss = \sum_{t} L_{rec}^{(t)} + L^{N(t)} + L^{A(t)}  \\
        &L^{N(t)} = \lambda^{N}L_{com}^{N{(t)}} + \mu^{N}L_{sep}^{N{(t)}} + \nu^{N}L_{tri}^{N{(t)}} \\
        &L^{A(t)} = \lambda^{A}L_{com}^{A{(t)}} + \mu^{A}L_{sep}^{A{(t)}} + \nu^{A}L_{tri}^{A{(t)}}
    \end{split}
    \end{equation}

    Following~\cite{liu2018future, gong2019memorizing, park2020learning}, we use Peak Signal to Noise Ratio (PSNR) between the predicted frame and its ground truth to quantify the reconstruction.
    \begin{equation} \label{equ:psnr}
        P(\hat{\mathbf{I}}, \textbf{I}) = 10log_{10} \frac{max(\hat{\mathbf{I}})}{||\hat{\mathbf{I}} - \mathbf{I}||_{2}^{2}/N}
    \end{equation}
    
    Although the abnormality memory is used to represent the frequent abnormality pattern, it cannot remember those unseen abnormal samples during training stage. It is inappropriate to absorb the separateness of abnormality memory into the abnormality score. Thus we compute the compactness using L2 distance between each normality query and the nearest items as follows:
    \begin{equation} \label{equ:compactnessscore}
        D(\mathbf{Q}^{N}) = \frac{1}{K} \sum_{k}^{K} ||\mathbf{q}_{k}^{N}-\mathbf{p}_{p}||_{2}^{2}
    \end{equation}
    However, it does not mean the abnormality branch contributes nothing to the final abnormality score. The abnormality branch generates a discriminative feature for both normal and abnormal data, which contributes to the reconstruction term in abnormality score. As our interpretation in the next section, the abnormality branch is essential for unknown anomaly types.
    
    The final abnormality score $\mathcal{S}$ for each video frame balances the two metrics by the parameter $\gamma$, as follows:
    \begin{equation} \label{equ:abnormalityscore}
        \mathcal{S} = \gamma (1-g(P(\hat{\mathbf{I}}, \mathbf{I}))) + (1-\gamma)g(D(\mathbf{Q}^{N}))
    \end{equation}
    where, $g(\cdot)$ is the min-max normalization over all frames of the whole video.

\subsection{An intuitive interpretation for DREAM}
\label{sec:modelinterp}
    In this part, we give an intuitive interpretation for our dual memory network in Fig.~\ref{fig:interpretation}, and show the capability of our network to deal with the unknown anomaly types. 
    
    The overall feature spaces are two $C$ dimension hyperspheres containing all unit queries and memory vectors. For both normality and abnormality memory space, their corresponding triplet losses are essentially leading to a closer distance among normal features and put aside abnormal features. As a result, we can obtain more recognizable feature spaces for both normality and abnormality. More importantly, the clusters of normal features and abnormal features are distant on the same hypersphere. 
    
    Compared with a single memory-based model, our dual memory can enhance the discriminative ability in normality and abnormality detection. An abnormal input will not only fail to be represented by normality memory items but also it can be represented by abnormality memory items as the abnormality memory is updated only by abnormal data. Similarly, a normal input can easily retrieve vectors from normality memory and fail to read any close abnormality memory vectors. Therefore, the reading features from both memory modules are effectively adding up the discrimination for the generator to give a more accurate score. Because the abnormality memories are separately updated in the abnormality space using only abnormal data, the abundant normal data will not overwhelm the abnormality patterns.
    
    Another problem is: What if an abnormal sample that has never been seen during the training phase appears? We address this issue from the following two aspects. 
    1) The queries of unknown abnormal samples are not close to any of the existing normality or abnormality memories, thus leading to a significant compactness loss to justify the final abnormality score. 
    2) The unknown abnormal data have weak responses to both normality and abnormality memory, so their reading features from both memory spaces will also be distinguishable compared with the normal data and the abnormal data learned by the model. This will further lead to a larger PSNR response for abnormality score.

    % In summary, our model benefits from a large number of normal data and a few abnormal data while has the ability to deal with unknown anomaly types.

\section{Experiments}
\label{sec:experiments}
\subsection{Datasets and evaluation metric}
    We evaluate our model on the four most popular open benchmark datasets and compare the results with several baselines. 1) The UCSD Ped1 dataset~\cite{li2013anomaly} contains 34 training and 36 testing videos with 40 uncommon events. All abnormal events are about objects such as scooters and bicycles. 2) The UCSD Ped2 dataset~\cite{li2013anomaly} includes 16 training and 12 testing videos with 12 irregular events, including riding a bike and driving a vehicle. 3) The CUHK Avenue dataset~\cite{lu2013abnormal} consists of 16 training and 21 testing videos with 47 abnormal events such as throwing stuff and rushing. 4) The ShanghaiTech dataset~\cite{luo2017revisit} contains 330 training and 107 testing videos of 13 scenes. It is the only dataset that consists of different backgrounds and the most challenging dataset for video anomaly detection. % 5) UCF Crime~\cite{sultani2018real}.

    Following the standard evaluation metric in the latest works~\cite{liu2018future, gong2019memorizing, park2020learning, liu2019margin}, we use the average frame-level area under curve (AUC) as the metric with varying threshold values for abnormality scores.

\subsection{Implementation details}
\label{sec:implementation_details}
    To predict a target frame, we use its previous consecutive four frames as the input. All frames are resized to $256 \times 256$ pixels and normalized to [-1, 1]. For both normality and abnormality branches, the height $H$, the width $W$ for query tensor, and channels $C$ for query and memory are set to 32, 32, 256, respectively. The item numbers $K$ for both normality and abnormality memories are set to 10. 
    UNet~\cite{ronneberger2015u} and ConvLSTM~\cite{liu2019margin}
    For most of the experiments, we use UNet~\cite{ronneberger2015u} as the backbone for the CNN feature extractor and frame generator. As ~\cite{liu2019margin} uses an ConvLSTM autoencoder as backbone, we also use ConvLSTM~\cite{liu2019margin} as backbone for a fair comparison. For both UNet and ConvLSTM, the normality and abnormality heads are split from the last two convolutional layers of encoder. Because a dual memory structure is used in our DREAM, to fairly compare against the baseline model with an equivalent amount of parameters, we reduce the channel numbers of both query and memory tensors by half (Table~\ref{table:parameters}).
    
    We use the Adam optimizer with $\beta_{1}$ = 0.9 and $\beta_{2}$ = 0.999, with a batch size of 4 for 60 epochs on all of UCSD Ped1, UCSD Ped2, CUHK Avenue, and ShanghaiTech, respectively. For all experiments, the learning rates are initialized to be 2e-5 and we decay them using the cosine annealing. We set $\lambda$ = 0.1, $\mu$ = 0.1, $\nu$ = 0.1 $\alpha$ = 1 and $\beta$ = 1 for both normality and abnormality branch. Specially, we set $\gamma$ = 0.6 for Ped1, Ped2, Avenue, and ShanghaiTech. It takes about 10, 3, 25 and 300 hours for training phase on UCSD Ped1, UCSD Ped2, CUHK Avenue, and ShanghaiTech, using two Nvidia RTX 2080 GPUs. The code is implemented in PyTorch~\cite{paszke2017automatic}. We will release the code soon.
    
    Our model looks more complex, but the time-consuming is closed to the other methods with the same backbone. The read and update operation consists of tensor multiplication, concatenation, softmax, and argmax. The scale of the tensor is not larger than ’batch x memory x channel’ and memory read and update operations are implemented using PyTorch with CUDA, so it is not timeconsuming. For one image of size 256x256, ours achieves 52 fps, while Mem-Guided~\cite{park2020learning} is 56 fps, MLEP~\cite{liu2019margin} is 65 fps, Frame-pred~\cite{gong2019memorizing} is 24 fps.
    
    \begin{table}
     \centering
     \caption{Parameters Number. Because we reduce the channel number by half, our model have the similar size as other memory based models. }
    \label{table:parameters}
    \begin{tabular}{ccccc}
    \toprule
        & \cite{park2020learning} & \cite{liu2019margin} & DREAM-UNet & DREAM-ConvLSTM  \\ 
    \midrule
    Size(M)   & 15.65  & 37.68  & 16.83 & 38.93 \\
    \bottomrule
    \end{tabular}
    \end{table}

\subsection{Comparison with other methods}
\label{sec:comparisonsota}
    Because only normal data are provided in training set in the standard training/testing split for all datasets, following ~\cite{liu2019margin}, we perform the standard $K$-fold cross-validation. In specific, we evenly split all original abnormal datasets into $K$ folds, choose one fold for each instance, and add it to the initial training set as the new training set. The rest of the abnormal data in the testing set are treated as the new testing set. In this experiment, we use $K$ = 10. We compare our models with baseline VAD models on UCSD Ped1, UCSD Ped2, CUHK Avenue, and ShanghaiTech (SHT for short) in Table~\ref{table:compare-sota}.
    
    Because all of the baseline methods except MLEP~\cite{liu2019margin} are evaluated on the original testing set, we re-run the results (denoted by *) of Frame-Pred~\cite{gong2019memorizing} and Mem-Guided~\cite{park2020learning} on new testing set. All published results of Frame-Pred, Mem-Guided and our reproduction on the new split testing dataset are provided in Table~\ref{table:compare-sota}. In our reproduction, for a fair comparison, we do not update the memory during testing as the code of Mem-Guided did.

    From Table~\ref{table:compare-sota}, we observe that our DREAM-UNet outperforms almost all the baselines, and Our DREAM-ConvLSTM also outperforms MLEP, which show the effectiveness of our dual module design.
    
    \begin{table}
     \caption{Quantitative comparison with baselines for anomaly detection. We measure the average AUC (\%) on UCSD Ped1, UCSD Ped2, CUHK Avenue, and ShanghaiTech. Numbers in bold indicate the best performance, and underscored ones are the second best. All the std is less than 0.01, and for space reason, we omit std in this table.}
    \label{table:compare-sota}
    \begin{tabular}{ccccc}
    \toprule
    AUC(\%)       & Ped1 & Ped2 & Avenue & ShanghaiTech  \\ 
    \midrule
    unmasking~\cite{tudor2017unmasking} & 68.4  & 82.2  & 80.6  & -     \\
    AMC~\cite{nguyen2019anomaly}        & -     & 96.2  & 86.9  & -     \\
    \addlinespace
    Conv-AE~\cite{hasan2016learning}    & 75.0  & 85.0  & 80.0  & 60.9  \\
    TSC~\cite{luo2017revisit}           & -     & 91.0  & 80.6  & 67.9  \\ 
    Stacked RNN~\cite{luo2017revisit}   & -     & 92.2  & 81.7  & 68.0  \\ 
    MemAE~\cite{gong2019memorizing}     & -     & 94.1  & 88.3  & 71.2  \\ 
    \addlinespace
    OGNet~\cite{zaheer2020old}          & -     & \underline{98.1}  & -     & -     \\
    Contrast~\cite{wang2020cluster}     & -     & -     & 87.0  & \underline{79.3}  \\
    Scene-Aware~\cite{sun2020scene}     & -     & -     & 89.6  & 74.7  \\
    
    \addlinespace
    Frame-Pred~\cite{liu2018future}     & 83.1  & 95.4  & 84.9  & 72.8  \\
    Frame-Pred*                         & 82.7  & 95.5  & 83.5  & 73.3  \\
    Mem-Guided~\cite{park2020learning}  & -     & 97.0  & 88.5  & 70.5  \\ 
    Mem-Guided*                         & 77.2  & 94.4  & 86.8  & 68.5  \\
    MLEP~\cite{liu2019margin}           & -     & -     & \underline{92.8}  & 76.8  \\
    \textbf{DREAM-UNet}                 & \underline{87.2}  & 97.8  & 91.2  & 72.7  \\
    \textbf{DREAM-ConvLSTM}             & \textbf{88.2} & \textbf{98.5} & \textbf{93.6} & \textbf{79.5} \\
    \bottomrule
    \end{tabular}
    \end{table}

\subsection{Ablation study}
    To further verify the effectiveness of our dual memory and discriminative branch, we show an ablation analysis of different components of our models in Table~\ref{table:ablation}. 
    
    For all the baselines except our DREAM-UNet (UNet+DualMem+Disc), we double the network channel numbers related to their feature space for a fair comparison of parameters. The UNet+Mem+Disc uses only a single branch in the memory module, and updates the normality and abnormality in the same space. UNet+Disc is an autoencoder without a memory module but with a discriminative network. The UNet+Mem is the reproduction for the Mem-Guided method, which does not use any abnormal training data. 
    
    The results again prove the effectiveness of the dual memory module for data imbalance without extra parameters. Comparing our DREAM(UNet+DualMem+Disc) with UNet+Mem+Disc, we can find that the dual memory design is more effective than the single memory design. Comparing all the methods with or without discriminator, we prove that the model with the discriminator using a few anomalies can get a better result.

    \begin{table}
     \caption{Ablation study}
    \label{table:ablation}
    %\begin{tabular}{m{89pt}m{15pt}m{15pt}m{26pt}m{26pt}}
    \begin{tabular}{ccccc}
    \toprule
    AUC(\%)         & Ped1 & Ped2 & Avenue & ShanghaiTech  \\ 
    \midrule
    UNet+Mem        & 77.2  & 94.4  & 86.8  & 68.5  \\
    UNet+Disc       & 81.9  & 95.1  & 86.1  & 69.4  \\
    UNet+Mem+Disc   & 83.9  & 96.0  & 87.7  & 71.0  \\
    \textbf{UNet+DualMem+Disc} & \textbf{87.2} & \textbf{97.8} & \textbf{91.2} & \textbf{72.7} \\
    \bottomrule
    \end{tabular}
    \end{table}

    \begin{figure}[t]
    \centering
    \subfigure[UCSD Ped1]{
        \label{fig:subfig:a_e} %% label for first subfigure
        \includegraphics[width=0.9\linewidth]{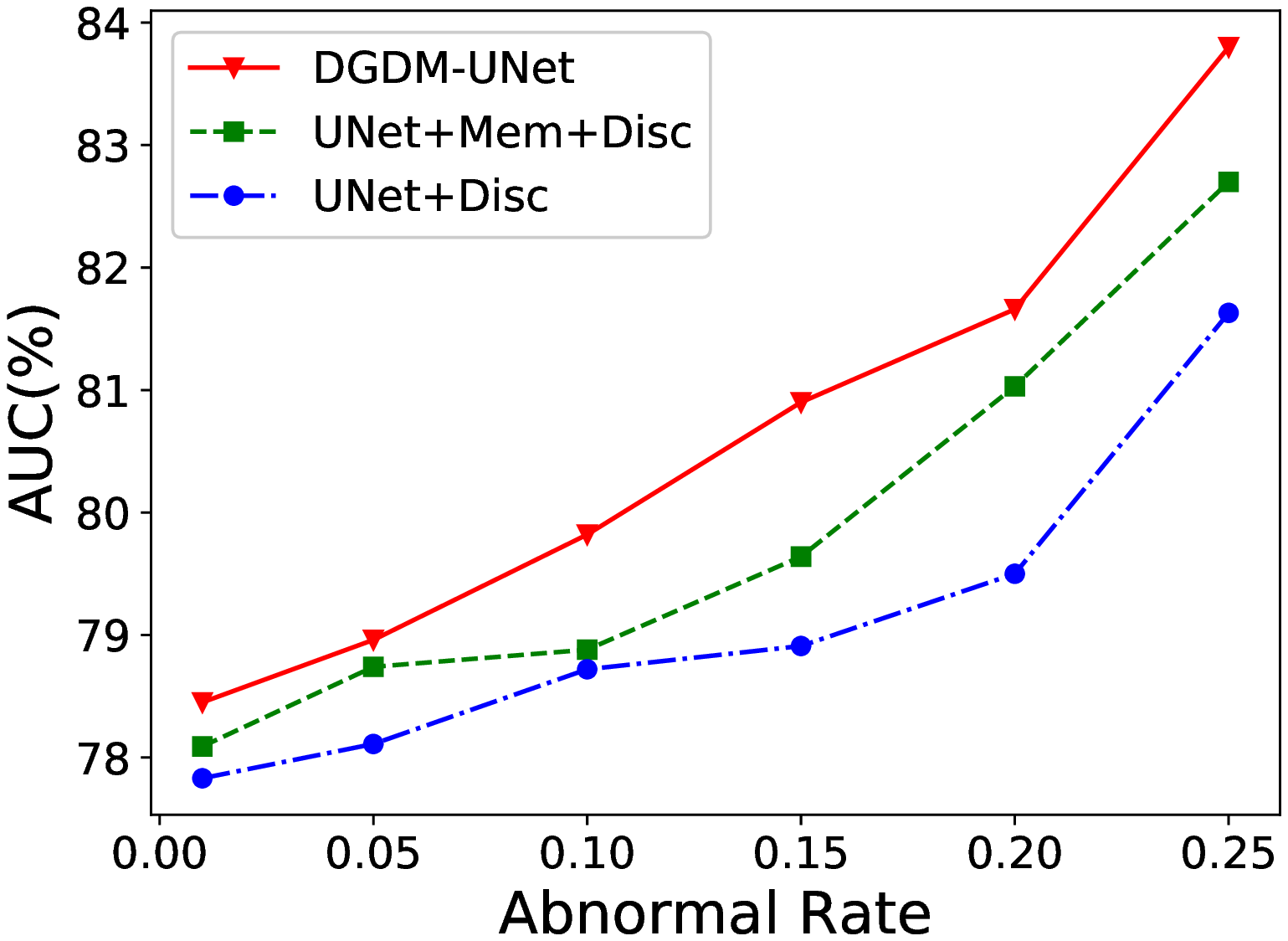}}
    \subfigure[UCSD Ped2]{
        \label{fig:subfig:b_e} %% label for second subfigure
        \includegraphics[width=0.9\linewidth]{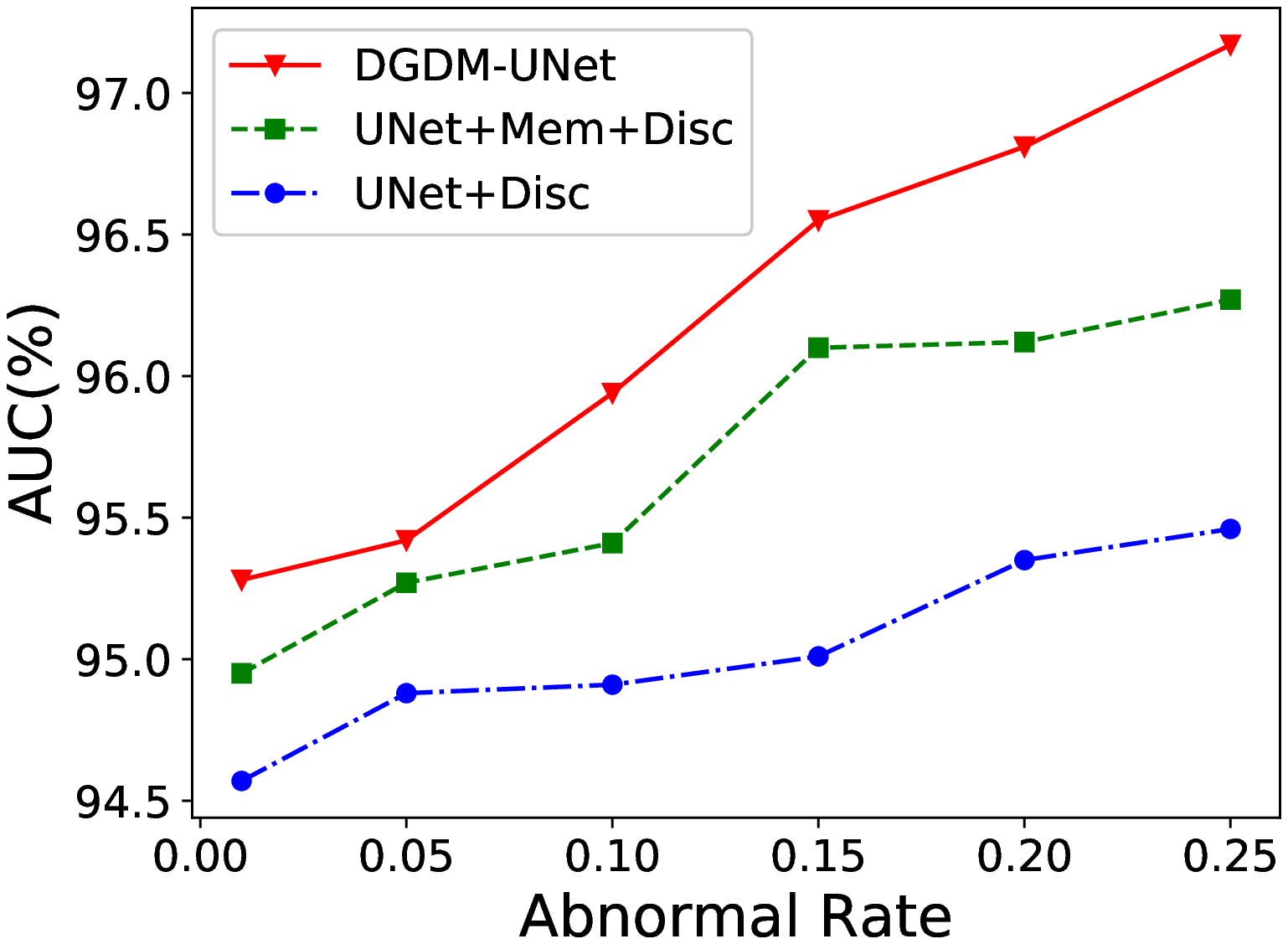}}
    \caption{Effect of different abnormal rates. Our DREAM is a robust model facing few anomalies, even if there is only 1\% abnormal data in the training set.}
    \label{fig:abnormalrate} %% label for entire figure
    \end{figure}
    
\subsection{Effect of different abnormal rates}
\label{sec:anomalyrate}
    In this part, we test our DREAM with the various abnormal rates (abnormal size / total training data size) for training data. In our experiment, the abnormal rates are set to be 1\%, 5\%, 10\%, 15\%, 20\%, and 25\%.
    
    Unlike the dataset splitting method in Sec.~\ref{sec:comparisonsota}, we randomly reserve half of the original abnormal data as the testing set, while we treat the other half as an abnormal training pool. According to the abnormal rate, we select the specific amount of abnormal samples from the pool and add them with normal training data to form the training set. Therefore, the new training set includes the original normal training data and the newly selected abnormal training data with a specific abnormal rate, while the testing set consists of the original normal testing data and reserved abnormal testing data.

    From Fig.~\ref{fig:subfig:a_e} and Fig.~\ref{fig:subfig:b_e}, our DREAM and other baselines work better with a higher abnormal rate. Our DREAM is a robust model facing few anomalies, even if there is only 1\% abnormal data in the training set.

\subsection{t-SNE visualization for feature space}
\begin{figure}[!ht]
    \centering
    \subfigure[Feature space for baseline: UNet+Mem+Disc]{
        \label{fig:subfig:a_t} %% label for first subfigure
        \includegraphics[width=0.6\linewidth]{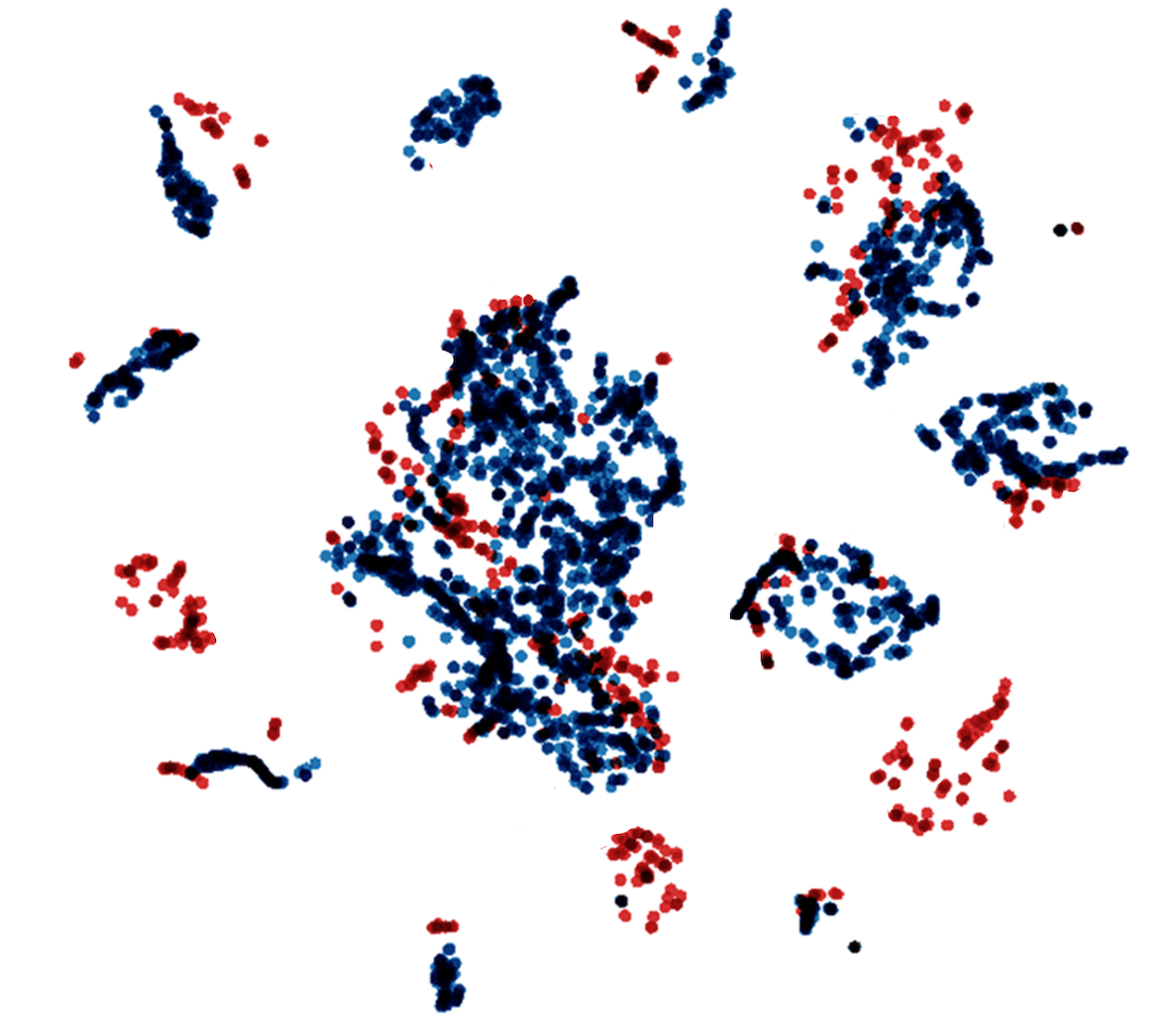}}
    \subfigure[Feature space for baseline: UNet+Disc]{
        \label{fig:subfig:b_t} %% label for first subfigure
        \includegraphics[width=0.6\linewidth]{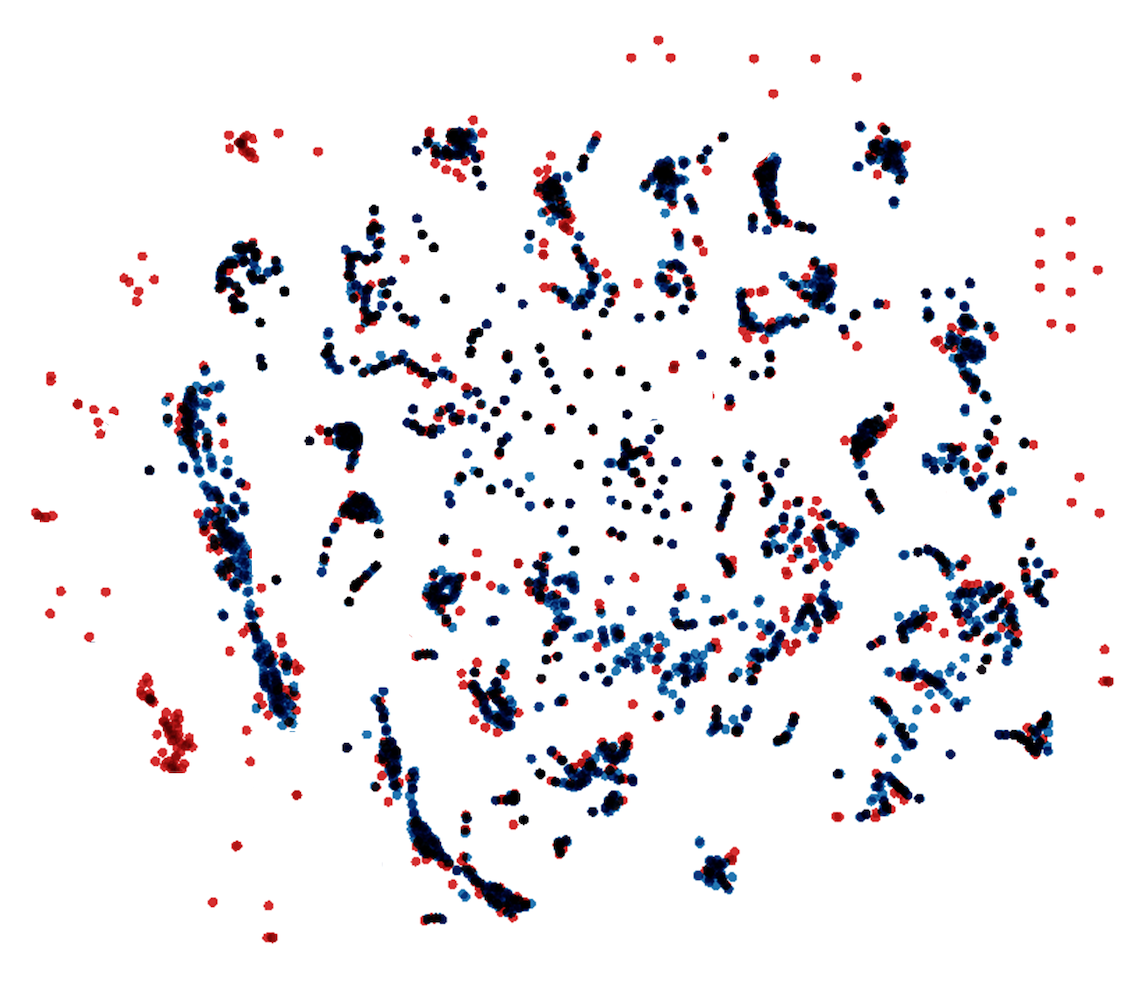}}
    \subfigure[Normality space of our model]{
        \label{fig:subfig:c_t} %% label for first subfigure
        \includegraphics[width=0.6\linewidth]{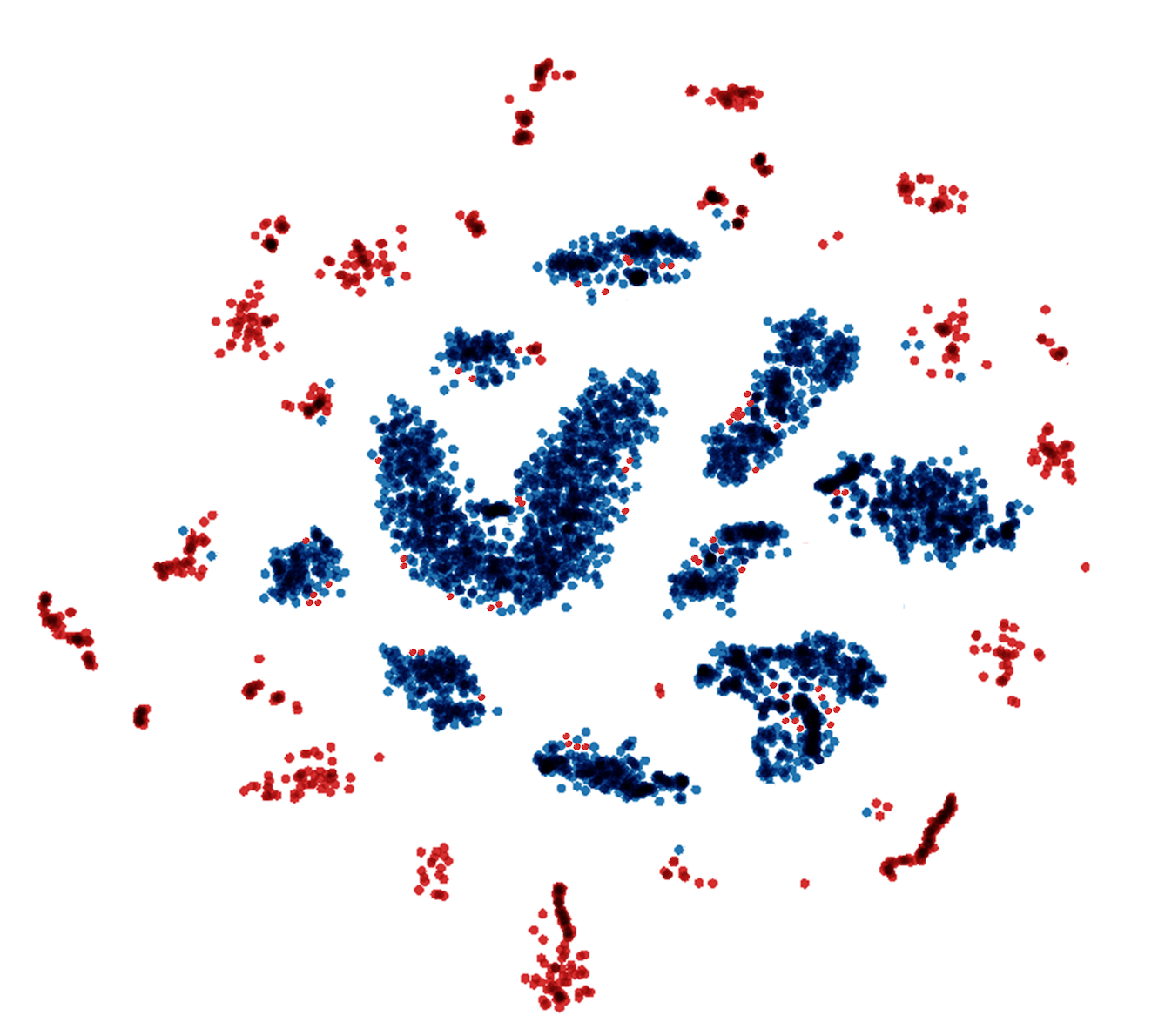}}
    \subfigure[Abnormality space of our model]{
        \label{fig:subfig:d_t} %% label for second subfigure
        \includegraphics[width=0.6\linewidth]{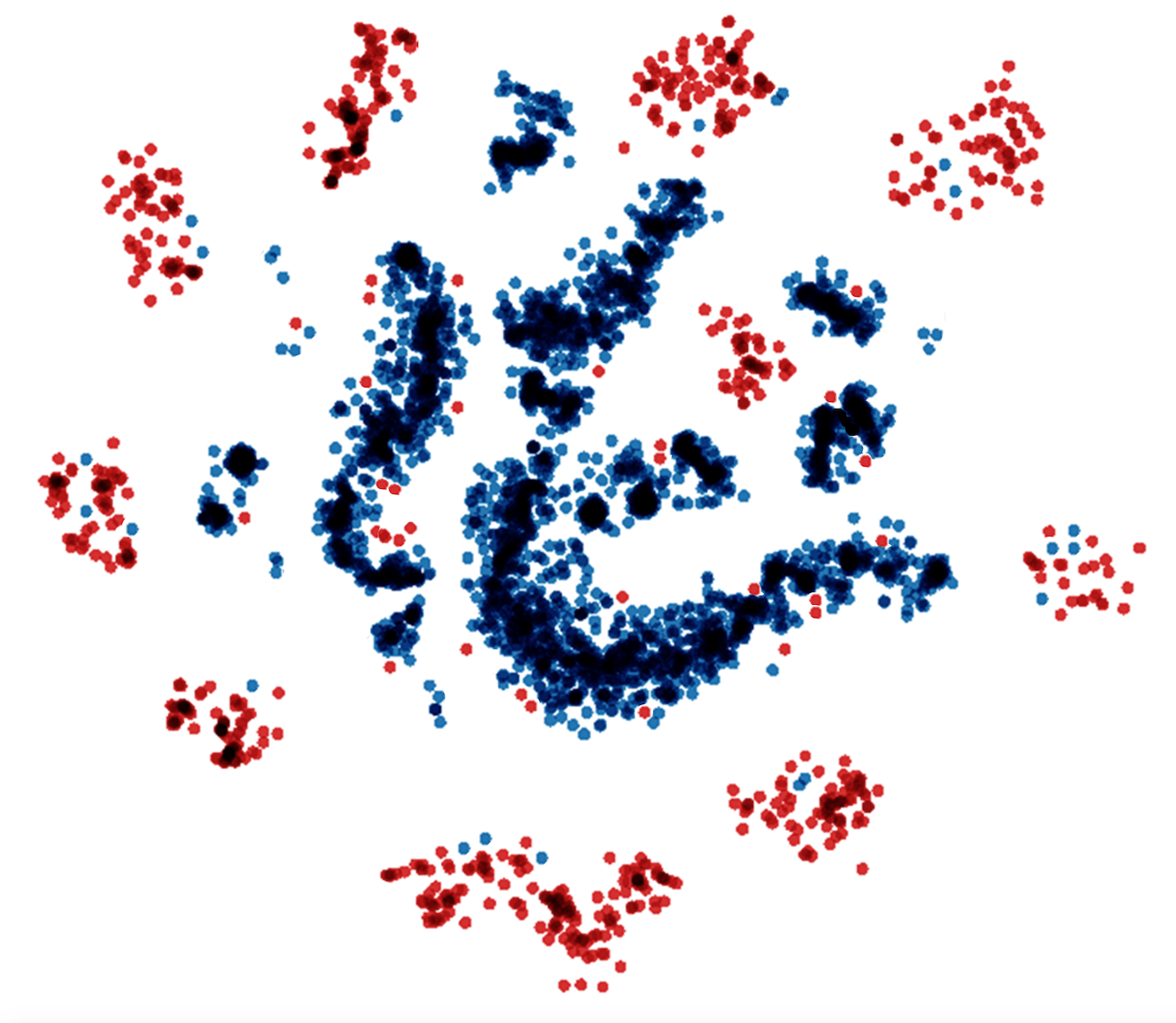}}

    \caption{t-SNE visualization for feature space on Avenue dataset. We represent the normal and abnormal features with blue and red points, respectively.}
    \label{fig:tsne} %% label for entire figure
    \end{figure}

    To prove our interpretation in Sec.~\ref{sec:modelinterp}, we use t-SNE~\cite{maaten2008visualizing} to visualize the feature space of our DREAM and compare with other baselines. Fig.~\ref{fig:subfig:a_t} shows that even with the supervised information, the UNet+Mem+Disc baseline cannot distinguish the normal and abnormal data perfectly. In this model, the normality memory and abnormality memory are updated in the same feature space, so it still suffers from the data imbalance problem. The small number of abnormal data is regarded as noise by the model and cannot learn a reasonable memory as we expect. In Fig.~\ref{fig:subfig:b_t}, without the memory module, feature distribution of the UNet+Disc is more random and scattered. From Fig.~\ref{fig:subfig:c_t} and~\ref{fig:subfig:d_t}, we can observe that, our DREAM learn a much more reasonable and distinct feature space. In the normality space, the normal features are grouped into several clusters because of the memory update. The abnormal features are distributed away from the normal clusters. Similarly, in the abnormal space, the abnormal feature is clustered, and the normal feature is forced to be closed to each other.
    
\subsection{Statistical Results}
To further quantitatively study the response of query vectors of normal/abnormal data to both normality/abnormality memories, we show the statistical results of the distance between query vectors and memories.

For each video clip $V_{t}=\{I_{t-T}, ...,I_{t-2}, I_{t-1}\}$, we collect all normality query vectors $\textbf{q}_{k}^{N}$ into set $U^{N}$ and all the abnormality query vectors $\textbf{q}_{k}^{A}$ into set $U^{A}$. Then we calculate the distance between normality query set $U^{N}$ and normality memory vectors set $M^{N}=\{\mathbf{p}_{m}^{N}\}_{m=1}^{K}$ (or abnormality memory vectors set $M^{A}=\{\mathbf{p}_{m}^{A}\}_{m=1}^{K}$) as Eq.~\eqref{equ:norm_distance0} (or Eq.~\eqref{equ:norm_distance1}). Similarly, the distance between abnormality query set $U^{A}$ and normality memory vectors set $M^{A}=\{\mathbf{p}_{m}^{N}\}_{m=1}^{K}$ (or abnormality memory vectors set $M^{A}=\{\mathbf{p}_{m}^{A}\}_{m=1}^{K}$) is defined in Eq.~\eqref{equ:abnorm_distance0} (or Eq.~\eqref{equ:abnorm_distance1}). $d(U^{N}, M^{N})$ describes whether the video clip $V_{t}$ is close to one of the normality memory. The smaller the value of $d(U^{N}, M^{N})$, the greater the possibility of $V_{t}$ to be normal. Similarly, an abnormal video clip is more possible to have a smaller $d(U^{A}, M^{A})$. 
    
    \begin{equation} 
    \label{equ:norm_distance0}
        d(U^{N}, M^{N}) = \underset{U^{N}}{avg}  \{ \underset{M^{N}}{min} \{ ||\textbf{q}_{k}^{N} - \mathbf{p}_{m}^{N}||_{2} \} \}
    \end{equation}
    
    \begin{equation} 
    \label{equ:norm_distance1}
        d(U^{N}, M^{A}) = \underset{U^{N}}{avg}  \{ \underset{M^{A}}{min} \{ ||\textbf{q}_{k}^{N} - \mathbf{p}_{m}^{A}||_{2} \} \}
    \end{equation}
    
    \begin{equation} 
    \label{equ:norm_rate}
        r_{normal} = \frac{\sum_{V_{t}} d(U^{N}, M^{N})} {\sum_{V_{t}} d(U^{N}, M^{A})}
    \end{equation}
    
    \begin{equation} 
    \label{equ:abnorm_distance0}
        d(U^{A}, M^{N}) = \underset{U^{A}}{avg}  \{ \underset{M^{N}}{min} \{ ||\textbf{q}_{k}^{A} - \mathbf{p}_{m}^{N}||_{2} \} \}
    \end{equation}
    
    \begin{equation} 
    \label{equ:abnorm_distance1}
        d(U^{A}, M^{A}) = \underset{U^{A}}{avg}  \{ \underset{M^{A}}{min} \{ ||\textbf{q}_{k}^{A} - \mathbf{p}_{m}^{A}||_{2} \} \}
    \end{equation}
    
    \begin{equation} 
    \label{equ:aabnorm_rate}
        r_{abnormal} = \frac{\sum_{V_{t}} d(U^{A}, M^{A})} {\sum_{V_{t}} d(U^{A}, M^{N})}
    \end{equation}

In this paper, we proposed to use the ratio of the sum of $d(U^{N}, M^{N})$ and $d(U^{N}, M^{A})$ to evaluate whether the normal features are closer to normality memories than abnormality memories. Also, we use the ratio of the sum of $d(U^{A}, M^{A})$ and $d(U^{A}, M^{N})$ to evaluate whether the abnormal features are closer to abnormality memories than normality memories. The results of the four datasets are shown in Table~\ref{table:statistical_results}. We can learn that our model can effectively distinguish the normal and abnormal features by the dual memory.

    \begin{table}
     \centering
     \caption{Statistical Results}
    \label{table:statistical_results}
    %\begin{tabular}{m{60pt}m{25pt}m{25pt}m{26pt}m{42pt}}
    \begin{tabular}{ccccc}
        \toprule
                   & Ped1 & Ped2 & Avenue & ShanghaiTech  \\ 
        \midrule
        $r_{normal}$     & 15.48  & 8.76  & 13.70  & 7.72  \\
        $r_{abnormal}$   & 12.72  & 9.97  & 10.91  & 8.83  \\
        \bottomrule
    \end{tabular}
    \end{table}

\subsection{Case study}
    In this section, we show the case study for our DREAM. The left image in each subfigure is the original image with at least one abnormal event. The right one is calculated as the absolute difference between the predicted frame and its ground truth. The cases include a man rushing, a girl stepping on the grass, paper falling, riding a bicycle, vehicle driving, and people fighting. The results show that our DREAM has a larger reconstruction loss for the abnormal event area. DREAM works well on both dynamic and static anomalies. 
        
    \begin{figure}[!ht]
    \centering
    \subfigure{
        \includegraphics[width=3.5cm]{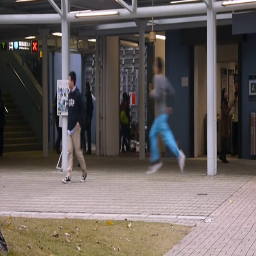}}
        \hspace{-0.23cm}
    \subfigure{
        \includegraphics[width=3.5cm]{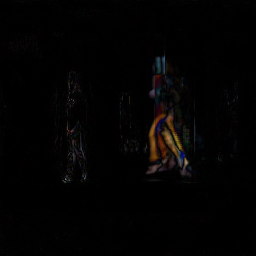}}
        \hspace{-0.17cm}
    \subfigure{
        \includegraphics[width=3.5cm]{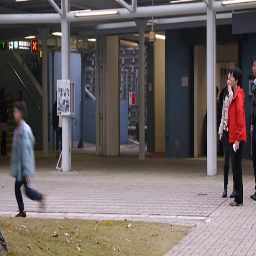}}
        \hspace{-0.23cm}
    \subfigure{
        \includegraphics[width=3.5cm]{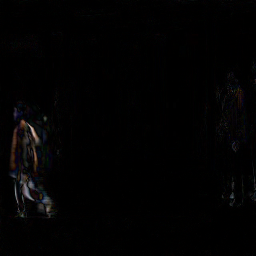}}
    \vspace{-0.25cm}
    
    \subfigure{
        \includegraphics[width=3.5cm]{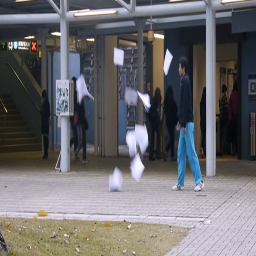}}
        \hspace{-0.23cm}
    \subfigure{
        \includegraphics[width=3.5cm]{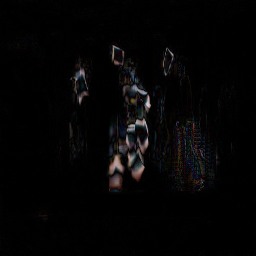}}
        \hspace{-0.17cm}
    \subfigure{
        \includegraphics[width=3.5cm]{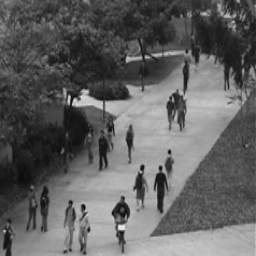}}
        \hspace{-0.23cm}
    \subfigure{
        \includegraphics[width=3.5cm]{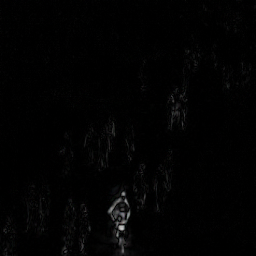}}
    \vspace{-0.25cm}
            
    \subfigure{
        \includegraphics[width=3.5cm]{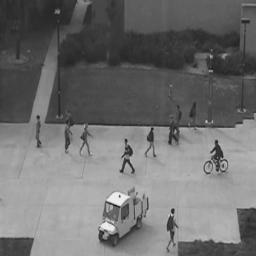}}
        \hspace{-0.23cm}
    \subfigure{
        \includegraphics[width=3.5cm]{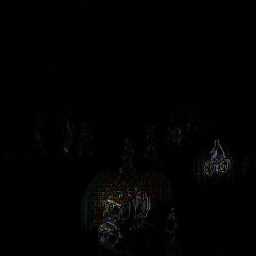}}
        \hspace{-0.17cm}
    \subfigure{
        \includegraphics[width=3.5cm]{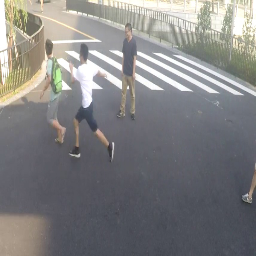}}
        \hspace{-0.23cm}
    \subfigure{
        \includegraphics[width=3.5cm]{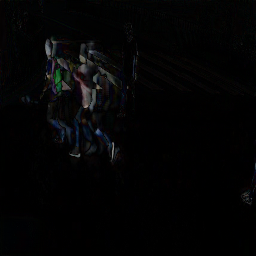}}
    \hspace{0.0cm}

    \caption{Case study. We show six cases, in which, the left one is the original image, the right is difference between the original image and our prediction.}
    \label{fig:case} %% label for entire figure
    \end{figure}

\section{Conclusion}
\label{sec:conclusion}
    We propose a Discriminative-Generative Dual Memory model for video anomaly detection. The DREAM uses two discriminative branches that utilize few anomalies to force the feature space to have an apparent distinction between normal features and observed abnormal features. We elaborately design a dual memory structure to separate the high-frequent normal data and the low frequent anomalous data so that the abnormal patterns will not be overwhelmed by the normal patterns. Our dual memory design can then solve the data imbalance problem. Extensive experiments show that our DREAM outperforms all of the other baseline methods in video anomaly detection. As far as we know, the dual memory is the first work that designs a particular module for video anomaly detection faced with data imbalance.

% or


%\appendices

%\section{}

\section*{Acknowledgment}

This work was supported in part by The National Key Research and Development Program of China (Grant Nos: 2018AAA0101400), in part by The National Nature Science Foundation of China (Grant Nos: 62036009, U1909203, 61936006), in part by the Alibaba-Zhejiang University Joint Institute of Frontier Technologies, in part by Innovation Capability Support Program of Shaanxi (Program No. 2021TD-05).

\ifCLASSOPTIONcaptionsoff
  \newpage
\fi



%\bibliographystyle{IEEEtran}

\bibliographystyle{IEEEtran}
\bibliography{reference}


\begin{IEEEbiography}
[{\includegraphics[width=1in,height=1.25in]{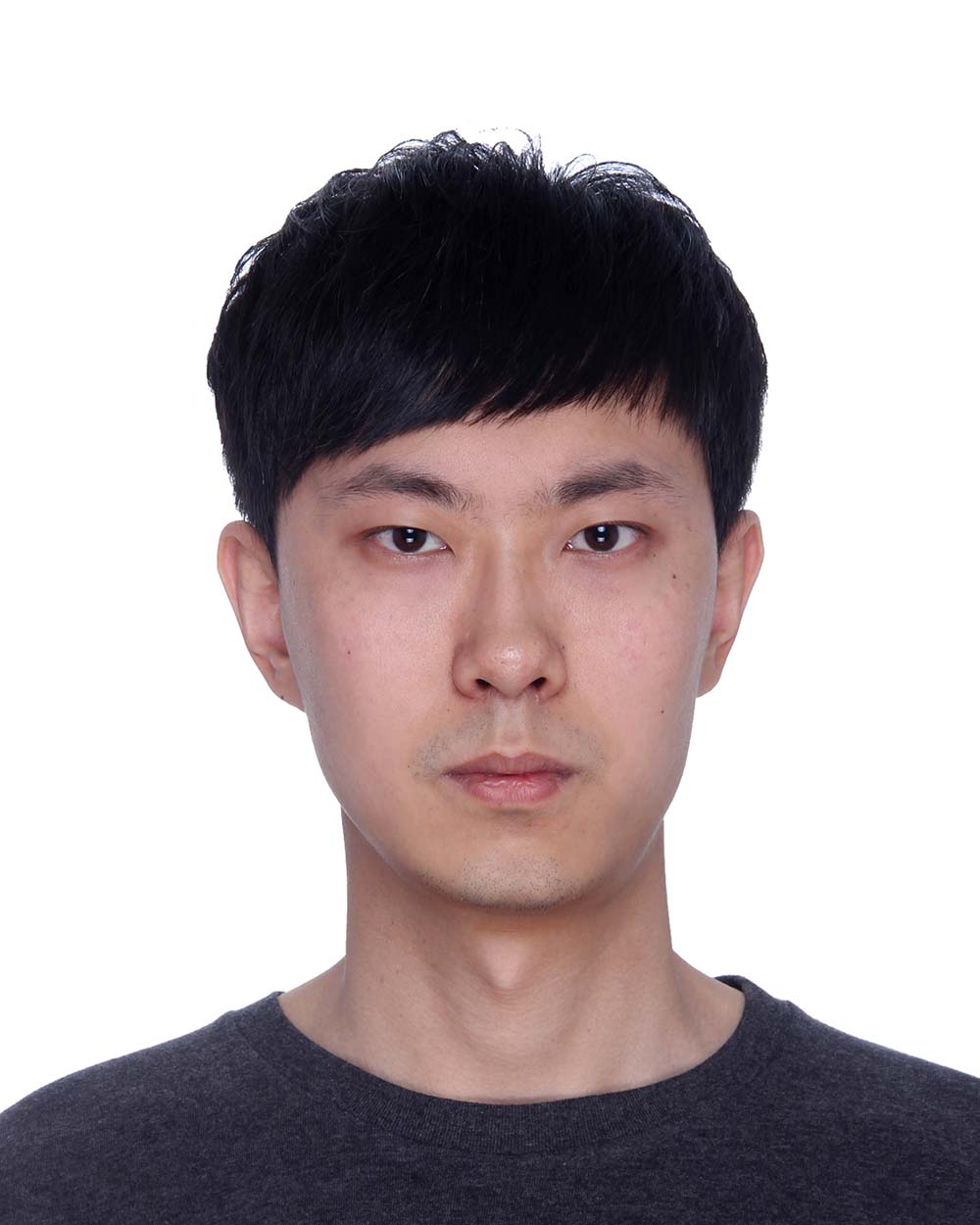}}]
{Xin Guo} received the B.S. degree in Mathematics and Applied Mathematics from Zhejiang University, Hangzhou, China, in 2015. He is now a Ph.D. candidate at the College of Computer Science and Technology, Zhejiang University, and also the State Key Laboratory of CAD \& CG. He is currently a research intern at Alibaba DAMO Academy. His research interests include computer vision and machine learning.
\end{IEEEbiography}

\begin{IEEEbiography}
[{\includegraphics[width=1in,height=1.25in]{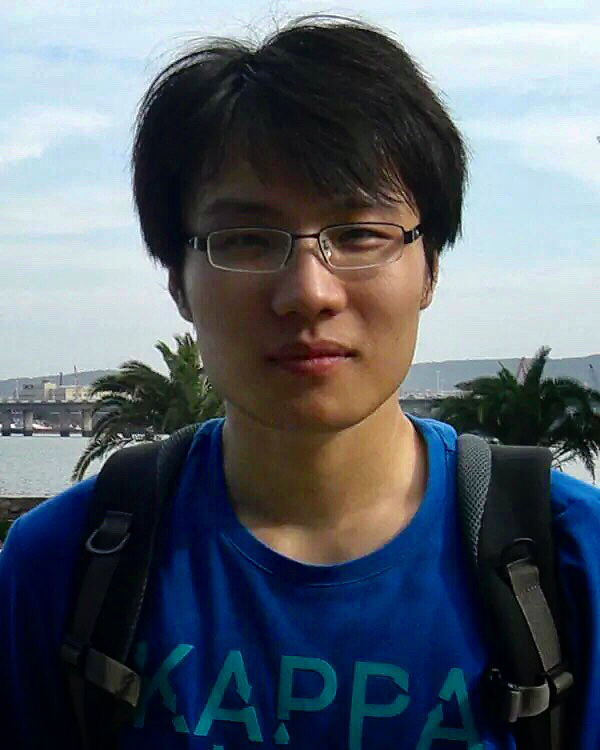}}]
{Zhongming Jin} is now a staff algorithm engineer at Alibaba DAMO Academy. Previously, he was a researcher at Baidu Research. He received his Ph.D. degree from Zhejiang University in Mar. 2015. His research interests include large scale machine learning and computer vision.
\end{IEEEbiography}

\begin{IEEEbiography}
[{\includegraphics[width=1in,height=1.25in]{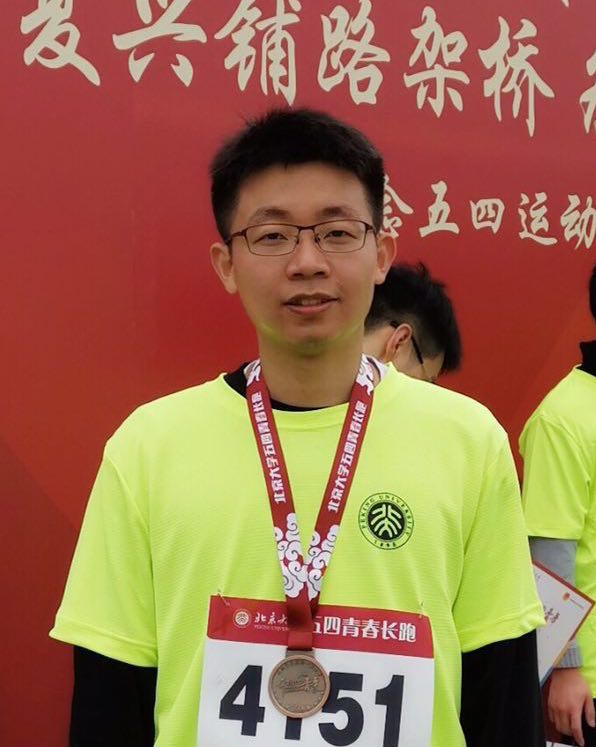}}]
{Chong Chen} is now a staff algorithm engineer at Alibaba DAMO Academy. He received the B.S. degree in Mathematics and Applied Mathematics from Peking University in June 2013. After that, he received the Ph.D. degree in Statistics from Peking University in June 2019. His research interests include Statistics,  machine learning and computer vision.
\end{IEEEbiography}

\begin{IEEEbiography}
[{\includegraphics[width=1in,height=1.25in]{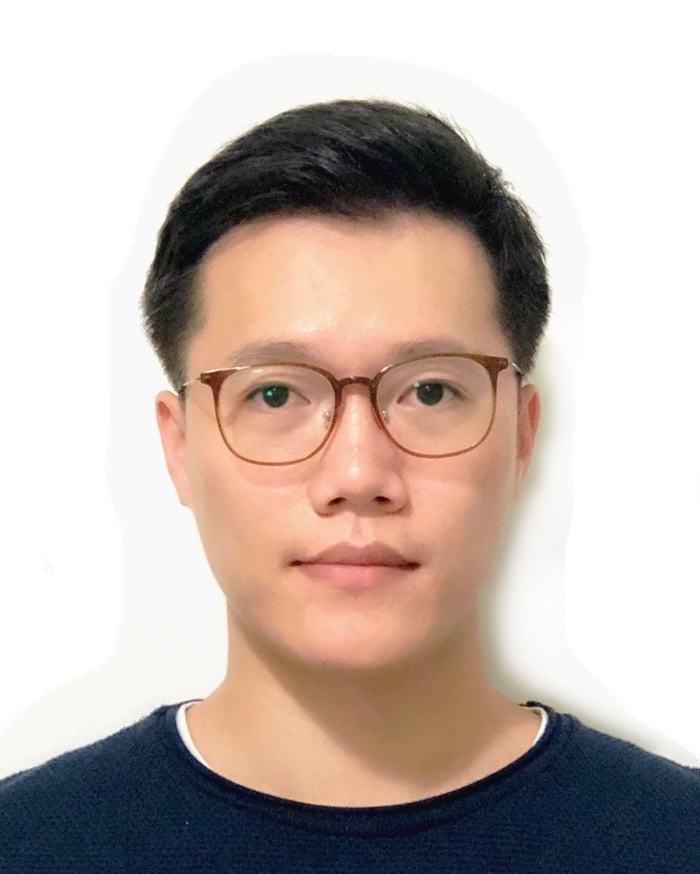}}]
{Helei Nie} received the B.S degree in Electrionic Engineering from Nanyang Techonological University (NTU), Singapore in 2015. He is currently a Ph.D. candidate at Interdisciplinary Graduate School NTU and also the Alibaba-NTU Singapore Joint Reserach Institute (JRI). His Research interests include machine learning and traffic data analysis. 
\end{IEEEbiography}

\begin{IEEEbiography}
[{\includegraphics[width=1in,height=1.25in]{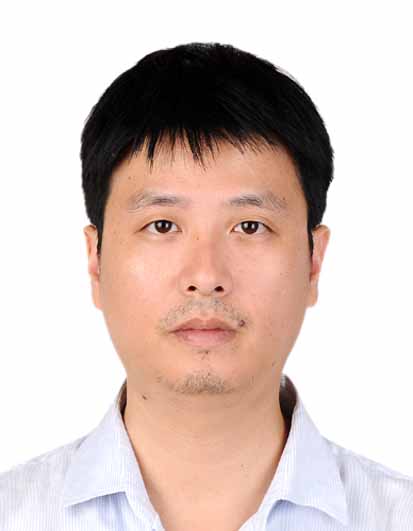}}]
{Jianqiang Huang} is a director of Alibaba DAMO Academy. He received the second prize of National Science and Technology Progress Award in 2010. His research interests focus on visual intelligence in the city brain project of Alibaba.
\end{IEEEbiography}

\begin{IEEEbiography}
[{\includegraphics[width=1in,height=1.25in]{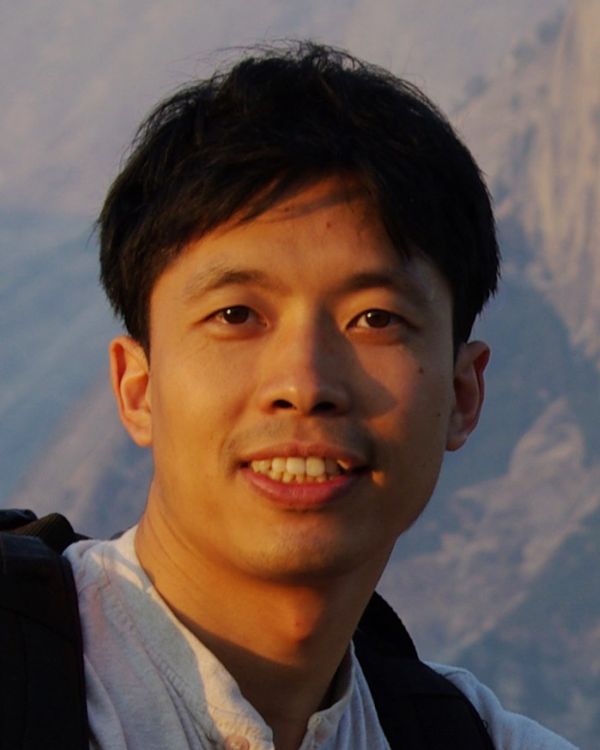}}]
{Deng Cai} is a professor in the State Key Lab of CAD\&CG, College of Computer Science at Zhejiang University, China. He received a Ph.D. degree in computer science from the University of Illinois at Urbana Champaign in 2009. His research interests include machine learning, data mining and information retrieval.
\end{IEEEbiography}

\begin{IEEEbiography}
[{\includegraphics[width=1in,height=1.25in]{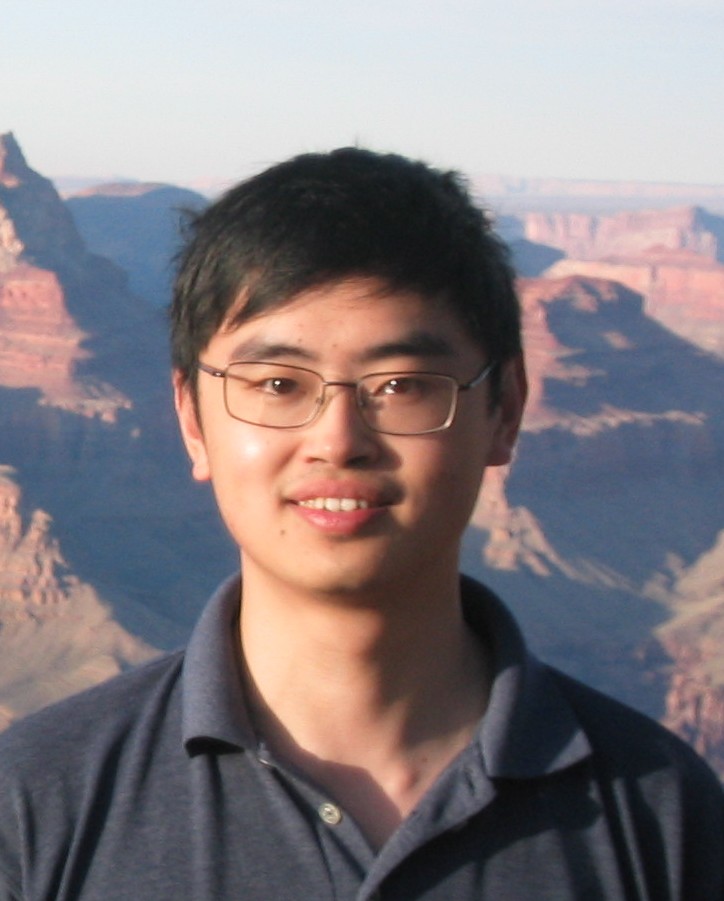}}]
{Xiaofei He} received a B.S. degree in Computer Science from Zhejiang University, China, in 2000 and a Ph.D. degree in Computer Science from the University of Chicago, in 2005. He is a Professor in the State Key Lab of CADCG at Zhejiang University, China. Prior to joining Zhejiang University, he was a Research Scientist at Yahoo! Research Labs, Burbank, CA. His research interests include machine learning, information retrieval,
and computer vision. He is a senior member of IEEE.
\end{IEEEbiography}

\begin{IEEEbiography}
[{\includegraphics[width=1in,height=1.25in]{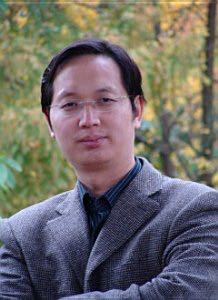}}]
{Xian-Sheng Hua} (F'16) received the B.S. and Ph.D. degrees in applied mathematics from Peking University, Beijing, in 1996 and 2001, respectively. In 2001, he joined Microsoft Research Asia as a Researcher
and has been a Senior Researcher of Microsoft Research
Redmond since 2013.
He became a Researcher and the Senior Director of the Alibaba Group in 2015.
He has authored or co-authored over 250 research papers and has filed over 90 patents. His research interests have been in the areas of multimedia search, advertising, understanding, and mining, and pattern
recognition and machine learning. 
He was honored as one of the recipients of MIT35.
He served as a Program Co-Chair for the IEEE ICME 2013, the ACM Multimedia 2012, and the IEEE ICME 2012, and on the Technical Directions Board of the IEEE Signal Processing Society. He is an ACM Distinguished Scientist and IEEE Fellow. 
\end{IEEEbiography}

\end{document}